This is the accepted version of the paper published in IEEE Access.
The final version is available at: https://doi.org/10.1109/ACCESS.2026.3665989

# Cluster Workload Allocation: Semantic Soft Affinity Using Natural Language Processing

**Leszek Sliwko[1], (Member, IEEE), Jolanta Mizeria-Pietraszko[2], (Member, IEEE)**
[1]Standard Chartered Bank, EC2V 5DD London, U.K.
[2]Department of Computer Science, Opole University of Technology, Opole, Poland

Corresponding author: Leszek Sliwko (e-mail: leszek.sliwko@sc.com).

This work was supported in part by the AI Research & Experiment department, Standard Chartered Bank.

**ABSTRACT** Cluster workload allocation often requires complex configurations, creating a usability gap. This paper introduces a semantic, intent-driven scheduling paradigm for cluster systems using Natural Language Processing. The system employs a Large Language Model (LLM) integrated via a Kubernetes scheduler extender to interpret natural language allocation hint annotations for soft affinity preferences. A prototype featuring a cluster state cache and an intent analyzer (using AWS Bedrock) was developed. Empirical evaluation demonstrated high LLM parsing accuracy (>95% Subset Accuracy on an evaluation ground-truth dataset) for top-tier models like Amazon Nova Pro/Premier and Mistral Pixtral Large, significantly outperforming a baseline engine. Scheduling quality tests across six scenarios showed the prototype achieved superior or equivalent placement compared to standard Kubernetes configurations, particularly excelling in complex and quantitative scenarios and handling conflicting soft preferences. The results validate using LLMs for accessible scheduling but highlight limitations like synchronous LLM latency, suggesting asynchronous processing for production readiness. This work confirms the viability of semantic soft affinity for simplifying workload orchestration and presents a proof-of-concept design.

**INDEX TERMS** Artificial Intelligence, Kubernetes, Load Balancing, Semantic Parsing, Soft-Affinity, Task Assignment

## I. INTRODUCTION

In any large-scale computing environment, from cloud data center to supercomputing cluster, the scheduler is one of the most critical components. Its fundamental purpose is to act as a primary coordinator, intelligently assigning user-submitted tasks, ranging from long-running services through short-lived data analysis jobs, to the most appropriate machines. The design of these schedulers has been a central research topic for decades, resulting in influential systems that define contemporary resource management. Representative examples include Google's Borg [9], which demonstrates how to orchestrate a vast and diverse collection of applications at unprecedented scale; Apache Hadoop YARN [8], which decouples resource management from job execution and has become the standard for big data workloads; and the SLURM Workload Manager, which dominates in High-Performance Computing (HPC) environments.

Schedulers typically make decisions based on two categories of rules: hard and soft ones. Hard rules, also known as constraints, represent absolute, non-negotiable requirements that a program needs to satisfy. For example, a graphics-intensive application may require a node equipped with a specific type of GPU. The scheduler first filters out all the machines that do not satisfy these constraints. If none of them qualify, the job is not scheduled, ensuring that applications never run in incompatible environments.

Once the scheduler identifies the machines satisfying all the hard rules, it applies the soft rules, or the preferences, to select the optimal candidate. These act as a guidance rather than some strict requirements, i.e. for preferring nodes that are lightly loaded or located near a data source in order to minimize latency. The scheduler assigns scores to the valid machines based on these preferences, optimizing for metrics such as performance or energy efficiency. Considerable research has focused on fairness and preference models, including Dominant Resource Fairness (DRF), which ensures equitable resource allocation among heterogeneous users [1].

These principles have been implemented in various systems. The Apache Mesos scheduler introduced a "resource offer" mechanism to manage constraints [3], while Firmament adopted a network-flow-based model to accelerate placement decisions [2]. Despite their significant architectural differences, these systems share a common limitation: they require operators to express scheduling goals in rigid, machine-readable syntax.

This leads to a significant semantic gap between human intent and machine-level configuration. A developer might express an intuitive request such as "Run this machine learning job on a powerful node, preferably one that isn't overloaded and is located in Europe." To achieve this in systems like Kubernetes, users must write complex YAML files specifying numerous fields and weighted values. This process is time-consuming, error-prone, and demands deep system expertise.

This paper introduces a new system designed to bridge that gap. The presented intelligent scheduler employs a Large Language Model (LLM) as a translation layer between natural-language user intent and the scheduler's structured directives. Developers can specify scheduling preferences in plain English (or any other supported language), and the model automatically generates the corresponding configuration, enabling the scheduler to make human-aligned placement decisions. The current prototype is intended as a research artifact to demonstrate the feasibility and is not production-ready.

The primary contribution is the design and implementation of a semantic, intent-driven scheduling paradigm for Kubernetes, which leverages an LLM to translate natural-language requirements into low-level scheduling logic. Specifically, the other contributions are:

- **Semantic Scheduling**: A shift from declarative, syntax-bound configuration (e.g., YAML with *nodeAffinity*) toward semantic, intent-based expressions. This reduces cognitive overhead for developers and supports more flexible workload specification.
- **Dynamic and Extensible Logic**: Traditional schedulers require writing and compiling new plug-ins in the Go programming language to incorporate custom logic. The presented approach allows new scheduling behaviors to be introduced dynamically by defining new Intent classes and updating the LLM prompt, with the aim of improving adaptability.
- **Real-Time Contextual Awareness**: By combining a live cluster state cache with the reasoning capabilities of an LLM, the system enables nuanced soft-affinity decisions (e.g., *prefer_colocate_same_deployment*, *spread_zones*) that go beyond static affinity and anti-affinity mechanisms.

The remainder of this paper is structured as follows: Section II reviews related work in cluster scheduling, highlighting existing approaches to affinity and resource management. Section III details the design and implementation of the prototype system and testbed environment, including the Kubernetes cluster simulation, the Cluster State Cache, the Score Extender Service, and the LLM-based Intent Analyzer module responsible for semantic parsing. Section IV details the mathematical formulas underlying the intent-driven scoring logic and explains the scalarization of the final score. Section V presents the empirical evaluation, focusing first on the accuracy of intent recognition using various LLMs against an evaluation dataset and then outlining the methodology for assessing scheduling efficiency and placement quality. Finally, Section VI concludes the paper by summarizing the key findings, discussing the limitations of the current prototype, and proposing directions for future research.

## II. RELATED WORKS

This section surveys foundational and contemporary research pertinent to the proposed system, structured across four principal domains. The first is AI in Cluster Scheduling, contrasting conventional policy-driven schedulers with ML and LLM-assisted approaches. The topic of semantic soft affinity intersects resource scheduling, ML-driven affinity prediction, and semantic analysis, drawing from workload trace analysis and AI-based scheduling. The second, Multi-Objective Scheduling (MOO), addresses the trade-offs in competing performance goals. The prototype's additive scoring model is a weighted-sum strategy. The third, Intent-Based Systems (IBS), focuses on bridging the "semantic gap" between high-level intents and low-level configurations. Lastly, Parsing and Natural Language Interfaces covers semantic parsing for converting linguistic input into machine-interpretable policies.

These research areas directly influence the system's architectural design. Insights from IBS and Natural Language Parsing (NLP) inform the Intent Analyzer. The foundations of MOO underpin the Score Extender Service, while advances in AI-assisted Scheduling guide the Cluster State Cache.

### *A. AI IN CLUSTER SCHEDULING*

Cluster scheduling is a central topic in distributed systems. Foundational systems like Google Borg [9], Apache Mesos [3], Quincy [4], and Firmament [2] established principles of fairness and resource sharing. Hadoop YARN [8] and SLURM are standards for big data and HPC. These schedulers rely on hard-coded rules and manual, declarative configuration.

Real-world traces, like Google's Cluster Data, offer foundations for understanding workload patterns. Alam et al. [38] clustered workloads by resource usage, revealing job distributions and symmetries. This supports semantic soft affinity by showing how clustering can identify resource preferences without hard constraints, enabling NLP-based grouping. Research conducted by Mishra et al. [42] also characterized Google workloads, noting elasticity and variability that inform soft affinity models.

ML is increasingly used to predict node-task affinities. Research by [40] proposed an ML ensemble to detect constraint operators, achieving 98% accuracy in affinity prediction. This predictive method could integrate NLP to parse metadata for softer, meaning-based affinities. Similarly, Mao et al. [6] introduced Decima, an RL framework that learns scheduling policies, improving job completion times. Decima's approach offers a basis for using NLP-derived semantic similarities.

Reviews highlight AI's role in overcoming traditional scheduling limitations. A study by [39] surveyed DL workload scheduling in GPU datacenters, noting techniques to optimize utilization, which underscores the need for semantic tools like NLP to handle diverse workloads. The survey conducted by [44] on AI-driven job scheduling examines ML, RL, and hybrid models for dynamic allocation. Zhou et al. [43] reviewed deep RL for cloud resource scheduling, emphasizing adaptive policies for NP-hard problems and suggesting semantic parsing extensions.

ML-assisted scheduling, using models to guide placement, is seen in systems like DeepRM [5] and Decima [6]. However, these approaches use fixed quantitative inputs, lack a natural language UI, and do not address the semantic gap between intent and configuration.

The emergence of LLMs inspired research into semantic, intent-driven configuration. OpenDevin [10] shows LLMs can interpret natural-language DevOps commands into configurations, using schema-guided extraction for deterministic parsing. Research by [7] also shows models can translate high-level goals into actions, though often outside latency-critical paths.

A growing body of research explores LLM-assisted cluster management, using natural-language understanding for configuration or categorization. Recent work by Jaillet et al. [11] focuses on scheduling LLM inference tasks with Key-Value cache constraints. These studies show an increasing intersection between natural-language reasoning and resource management. The present work extends this by focusing on semantic intent translation for soft-affinity scheduling, using an LLM with a live scheduler extender and a reproducible testbed. It contributes to integrating natural-language interfaces into infrastructure orchestration, emphasizing reproducibility, determinism, and quantitative evaluation.

### B. MULTI-OBJECTIVE SCHEDULING

The task of cluster scheduling is fundamentally a MOO problem, requiring trade-offs between conflicting objectives [33]. For example, maximizing fault tolerance by spreading pods (a *spread_zones* intent) conflicts with minimizing network latency by co-locating them on the same nodes (a *prefer_colocate_same_deployment* intent). Other conflicts include balancing resource utilization versus consolidating workloads for energy savings [31], or minimizing cost versus minimizing job completion time [34].

Formally, these problems seek Pareto optimal solutions, where no objective can be improved without degrading another. Research employs two main strategies: evolutionary algorithms (e.g., Non-dominated Sorting Genetic Algorithm II) to find the Pareto front, or scalarization. Scalarization, the more common approach, converts the multi-objective problem into a single-objective one, often via a weighted-sum method [34].

A key challenge is that presenting an entire Pareto front is impractical for real-time decisions. This led to preference-based optimization, using high-level user preferences to guide the selection of a single best-fit solution. User preferences define the weights or constraints for the scalarization function [32]. The allocation hint proposed in this work serves this function; it is a mechanism for capturing unstructured preferences and translating them, via an LLM, into the quantitative weights for the scheduler's multi-objective cost function.

### C. INTENT-BASED SYSTEMS

Research in IBS addresses the “semantic gap” between high-level human goals and low-level system configuration [13]. The premise is that operators should specify what they want (e.g., “ensure low latency”) rather than how to achieve it (e.g., setting specific routing rules) [14]. This declarative, policy-driven model aims to replace imperative, error-prone manual configurations, reducing human error and cognitive overhead [16]. The core challenge is translating these abstract, human-centric goals into precise, machine-executable instructions.

An IBS typically operates as a continuous, closed-loop lifecycle [21]. This loop includes: intent ingestion (capturing the objective); translation (converting the intent into formal policies); validation (ensuring policies are achievable and non-conflicting); actuation (orchestrating policies via controllers); and assurance (continuously monitoring the state to ensure the intent is met) [18]. If intent drift is detected, the system takes corrective action, completing the autonomous loop.

The intent ingestion and translation phases are most relevant to this work. Traditionally, these use structured inputs like GUIs or, more commonly, Domain-Specific Languages (DSLs) [17]. DSLs allow operators to define complex intents formally [20]. While an improvement, DSLs still require specialized training and present a steep learning curve, creating a usability gap.

IBS principles are most applied in Intent-Based Networking (IBN), particularly in SDN and managing 5G [24] and 6G [18] networks. Intents are used for traffic engineering, network slicing, and enforcing dynamic QoS [21]. The paradigm is also used in multi-cloud deployment [22] and defining dynamic security policies [19].

To address the usability limitations of formal DSLs, research has emerged to use natural language as the primary intent input. Early work explored NLP for translating user commands into network control actions [23] or verifiable policies [24]. The present work builds on this trajectory, proposing that modern LLMs can serve as a more powerful “universal translator” for intent. By accepting unstructured allocation hints, this approach aims to eliminate specialized syntax, making intent-based scheduling accessible without the steep learning curve of traditional IBS.

### D. PARSING AND NATURAL LANGUAGE INTERFACES

The Intent Analyzer's core function—translating an unstructured hint into a structured JSON object—is a semantic parsing challenge. This is extensively studied in Natural Language Interfaces to Databases (NLIDB), which aim to translate plain text into a formal query language like SQL [12]. While traditional NLIDB systems struggled with ambiguity, advances in deep learning and LLMs have significantly enhanced semantic parser performance [15]. This research applies these techniques to cluster scheduling, mapping text to a scheduling policy rather than a database query.

To enable semantic soft affinity, NLP-driven clustering of textual workload descriptions (e.g., job requirements) is crucial. Veremyev et al. [41] apply graph-based clustering to semantic spaces from word embeddings, revealing dense communities that capture contextual meanings—directly applicable to grouping similar jobs for soft allocation. This aligns with broader NLP applications in resource management, though explicit integration with cluster scheduling remains underexplored.

The advent of deep learning introduced new text-to-SQL architectures. Early sequence-to-sequence models struggled with rigid SQL syntax [30]. This led to sketch-based methods, where models like SQLNet learned to fill slots in a predefined SQL sketch, improving accuracy [29]. Pre-trained transformers like BERT enabled models such as SQLova to consider both the question's context and the database schema's structure [27]. A critical sub-task is schema linking: identifying relevant database tables and columns for a question. In large databases, including the entire schema is inefficient; effective schema linking prunes the schema to only relevant parts. However, incorrectly omitting a required table can impede correct query generation [26].

More recently, general-purpose LLMs have been applied to the NLIDB problem. Unlike earlier models needing extensive fine-tuning, LLMs can often perform text-to-SQL translation in a zero-shot or few-shot setting, generating correct queries for new schemas with little to no task-specific training [28]. This approach bypasses the costly data annotation process.

Despite progress, significant challenges remain [25]. Models still struggle with domain adaptation (handling specialized terminology) and logic grounding (understanding abstract concepts). Generating highly complex SQL queries remains a research frontier. These works form a strong foundation, with gaps in NLP-specific semantic affinity offering opportunities for novel contributions. Future research could hybridize RL schedulers with NLP models for intent-based soft affinities.

## III. PROTOTYPE AND TESTBED DESIGN

To empirically evaluate the performance and efficacy of an intent-based scheduling paradigm, a robust and reproducible testbed is required. This section details the design and implementation of such an environment. The testbed is built around a multi-node Kubernetes cluster simulated using Minikube, alongside a custom scheduler extender. The extender intercepts scheduling requests and translates high-level, natural-language intents into concrete node-scoring decisions.

The environment (see Figure 1 for detail) is designed as a self-contained representation of a topology-aware data center, enabling the evaluation of complex scheduling policies such as proximity preference, resource affinity, and workload spreading. The core components include:

- **Local Kubernetes Cluster**: A multi-node Minikube cluster deployed via Minikube with the Docker driver, with nodes labeled to simulate a multi-level physical topology.
- **Cluster State Cache**: A real-time, thread-safe subsystem that maintains an up-to-date in-memory representation of cluster state.
- **Score Extender Service**: A stateful scheduler extender microservice where the custom node-scoring logic resides.
- **Intent Analyzer**: An NLP module leveraging AWS Bedrock to access an LLM to translate user-provided hints into structured scheduling intents.

Each of these modules, along with the testbed environment as a whole, is described in detail in the following sections.

### A. LOCAL KUBERNETES CLUSTER

The foundation of the experimental testbed is a local Kubernetes version 1.31.4 cluster deployed using Minikube, configured with the Docker driver for node virtualization. This setup provides a fully contained, reproducible environment suitable for rapid iteration and controlled experimentation. Figure 2 shows how the cluster is initialized with nine nodes: one dedicated control-plane node (*minikube*) and eight worker nodes (*minikube-m02* through *minikube-m09*). This topology offers sufficient scale to evaluate scheduling behaviors across multiple failure domains while maintaining the manageability required for iterative testing.

The Minikube testbed is initialized using the Docker driver, providing lightweight, containerized node virtualization instead of full virtual machines. This approach offers several advantages: it eliminates hypervisor configuration, reduces resource overhead compared to VM-based drivers, and shares the host’s networking stack for fast pod-scheduler communication. These containerized nodes can be managed via the Docker dashboard and programmatically scripted (e.g., labeled, tainted), which makes the testbed highly reproducible. This setup simplifies dependency management by using the host's OS and runtime, providing a flexible platform for experimenting with topology-aware scheduling, custom extenders, and LLM-driven intent analysis without requiring a production cluster.

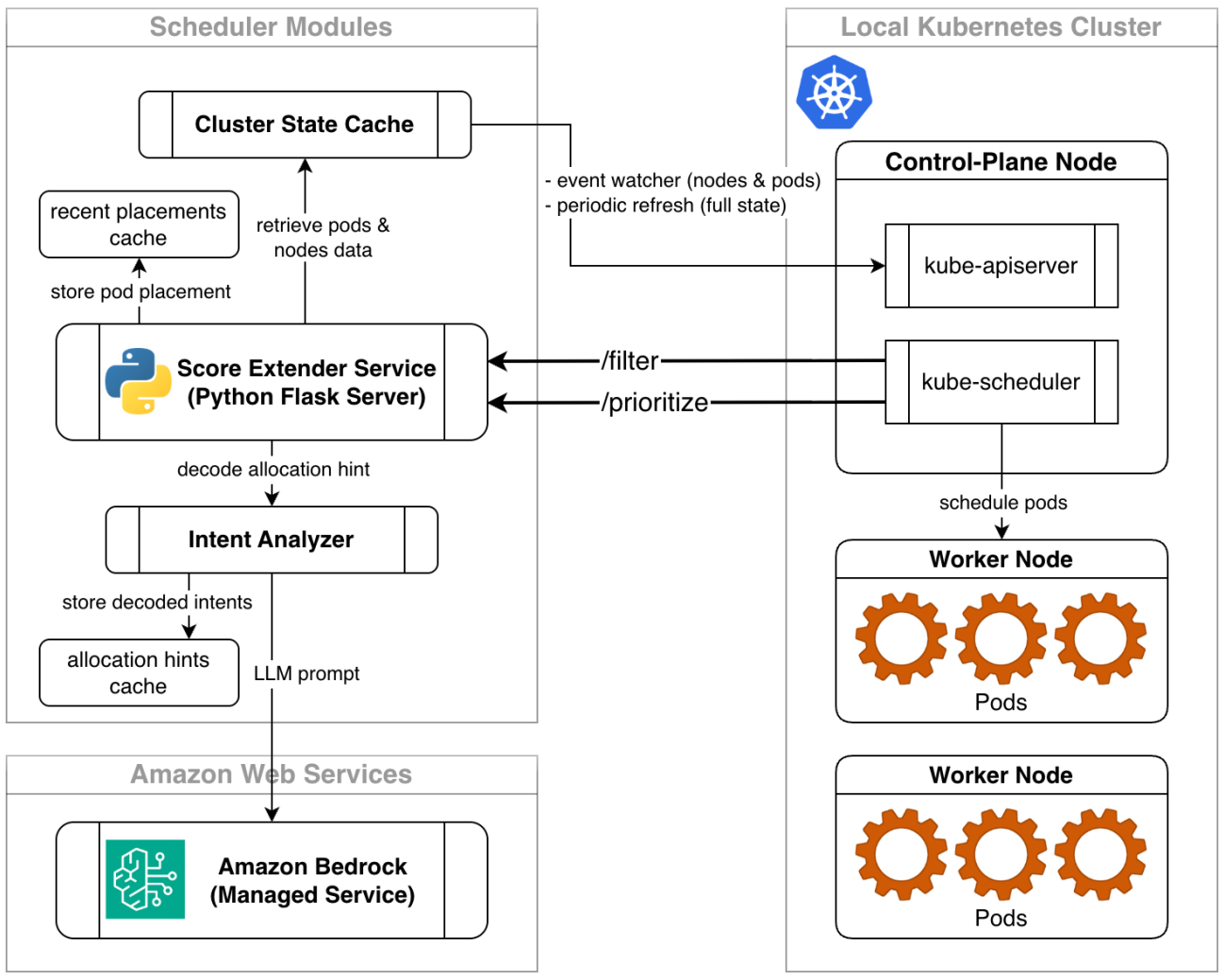

**FIGURE 1. Prototype system architecture, illustrating the interaction between the main components: the Local Kubernetes Cluster (Minikube), Cluster State Cache, Score Extender Service, Intent Analyzer, and AWS Bedrock.**

A critical feature of the testbed is its emulation of a physical network topology, necessary for validating topology-aware scheduling intents such as *prefer_colocate_same_deployment*, *spread_zones*, and *prefer_nearby_nodes_same_deployment*. Immediately after cluster initialization, a setup script programmatically labels the worker nodes to define a hierarchical topology consistent with standard Kubernetes conventions. The hierarchy uses the following topology keys:

- **Region** (*topology.kubernetes.io/region*): All nodes share a single region label, us-east-1.
- **Zone** (*topology.kubernetes.io/zone*): Two simulated availability zones, us-east-1a and us-east-1b, are assigned across the worker nodes.
- **Rack** (*topology.kubernetes.io/rack*): Five logical racks (rack-1 through rack-5) are distributed across the zones, with two nodes per racks 1, 3, and 4, and one node per rack 2 and 5.

TABLE I
EVALUATION CLUSTER TOPOLOGY

| Node | Region | Zone | Rack |
|---|---|---|---|
| *minikube-m02* | us-east-1 | us-east-1a | rack-1 |
| *minikube-m03* | us-east-1 | us-east-1a | rack-1 |
| *minikube-m04* | us-east-1 | us-east-1b | rack-2 |
| *minikube-m05* | us-east-1 | us-east-1b | rack-3 |
| *minikube-m06* | us-east-1 | us-east-1b | rack-3 |
| *minikube-m07* | us-east-1 | us-east-1b | rack-4 |
| *minikube-m08* | us-east-1 | us-east-1b | rack-4 |
| *minikube-m09* | us-east-1 | us-east-1b | rack-5 |

Table I details the topology structure used in the further evaluation scenarios. The labeling scheme enables the prototype scheduler to interpret placement hints and make decisions based on synthetic but realistic topology metadata, closely mimicking the conditions of a multi-zone, multi-rack data center environment. To ensure isolation of control-plane operations, the control-plane node is tainted with *node-role.kubernetes.io/control-plane=:NoSchedule*, preventing the placement of user workloads and reserving the node exclusively for critical cluster management services such as *etcd*, *kube-apiserver*, and *kube-scheduler*.

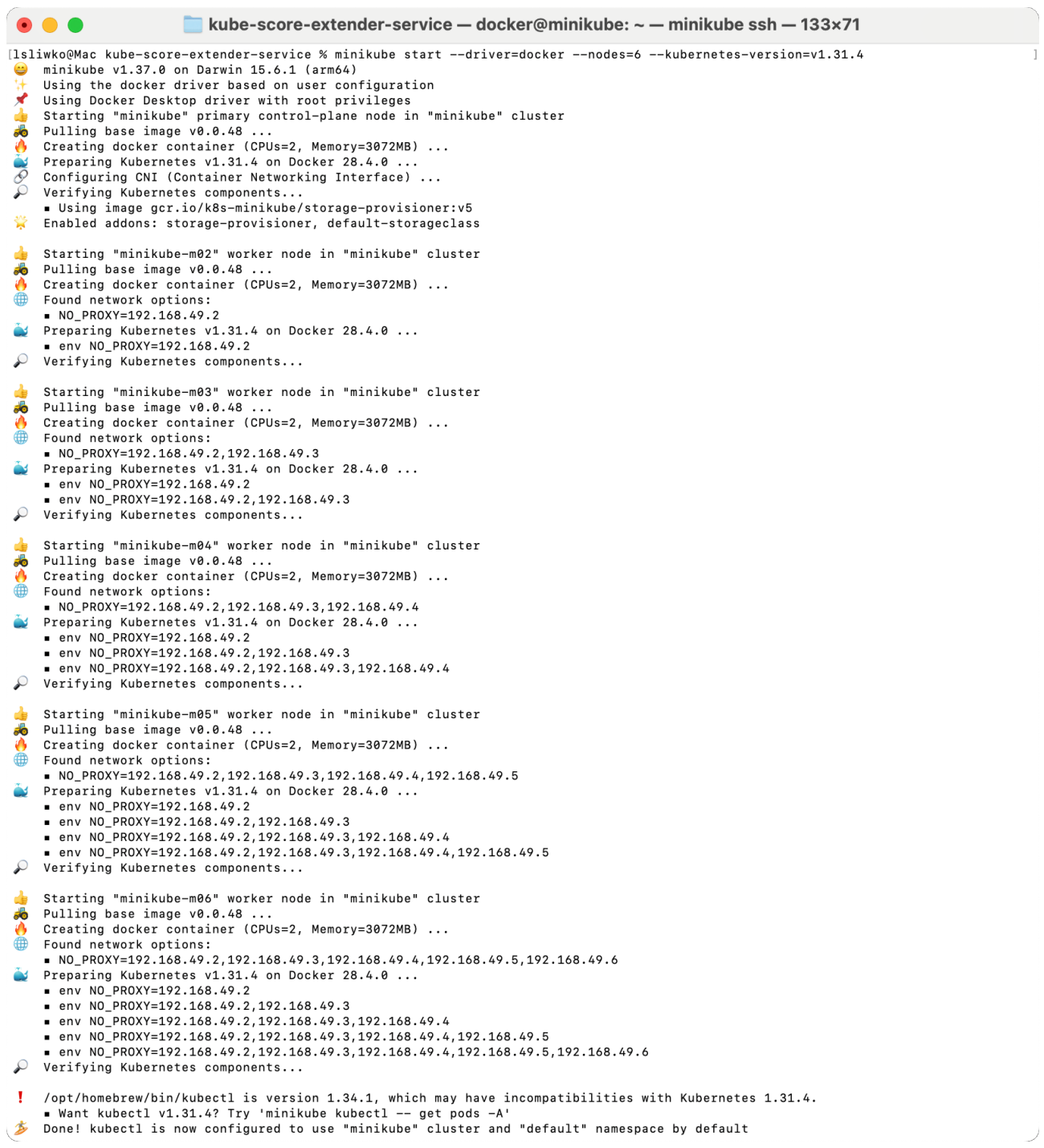

**FIGURE 2. Minikube cluster startup sequence (Kubernetes version 1.31.4). The cluster consists of nine nodes: one control-plane node (*minikube*) and eight worker nodes (*minikube-m02* through *m09*).**

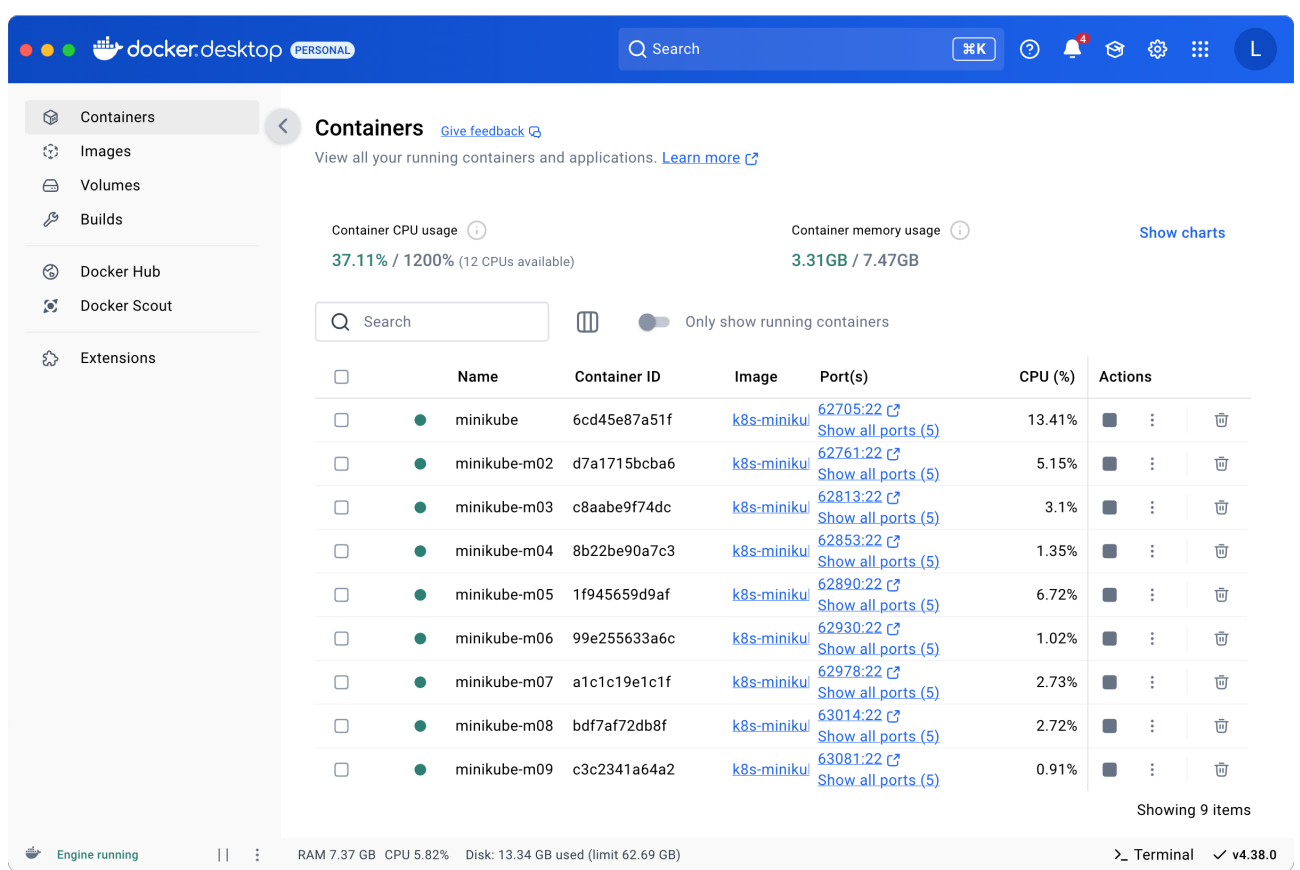

**FIGURE 3. Minikube nodes displayed in the Docker Desktop dashboard. This GUI tool provided a convenient way to monitor the state of the Kubernetes cluster nodes and manage associated persisted state (Volumes).**

The Score Extender Service, which encapsulates the semantic intent-based scheduling logic, is integrated directly into the Kubernetes control-plane via a custom scheduler configuration file. This configuration modifies the kube-scheduler manifest to include the custom extender and adjust its scheduling behavior. Two configuration aspects are especially important:

- **Extender Weight**: The extender's weight in the scheduler configuration is set to 100, the maximum possible value. This ensures that the extender's node scores (ranging from 0-100) dominate those from Kubernetes' native scoring plugins, giving the prototype primary control over node selection.
- **Disabling Default Scoring Plugins**: To isolate the effects of intent-based scheduling, several built-in Kubernetes scoring plugins are explicitly disabled, including *NodeResourcesFit*, *TaintToleration*, *InterPodAffinity*, and *DefaultPodTopologySpread*. Disabling these ensures that all node scoring and prioritization decisions originate from the custom NLP-driven scheduler extender, allowing for accurate evaluation of its independent behavior.

This Minikube-based testbed thus provides a controlled, topology-aware sandbox that replicates key aspects of real-world cluster scheduling while remaining lightweight and reproducible. It enables precise measurement of scheduling latency, accuracy, and correctness under various intent configurations, while isolating the impact of the LLM-driven logic from external Kubernetes behaviors. Notably, in Kubernetes, the scheduler evaluates resource availability based on requested values rather than real-time monitoring of current CPU load. Pod specifications include requests and limits for CPU and memory, which are used during the filtering phase to ensure a node has enough unallocated capacity to satisfy a pod's requirements. During this initial filtering phase (*/filter*), native Kubernetes plugins like *NodeResourcesFit* remain active to exclude nodes that cannot support the pod's mandatory resource needs. Only after nodes have passed these hard-rule filters are they sent to the prioritization phase (*/prioritize*), where the prototype's custom extender applies its additive scoring logic.

### B. CLUSTER STATE CACHE

The scheduler extender must make rapid decisions based on the current state of the entire cluster. Continuously querying the Kubernetes API server for each */prioritize* request is computationally infeasible, as it would introduce prohibitive latency and place an unsustainable load on the API server. To address this, a thread-safe state management subsystem the Cluster State Cache was implemented.

The Cluster State Cache is designed as a distributed state synchronization mechanism that balances low-latency responsiveness with eventual consistency. This service maintains a continuously updated, in-memory representation of all nodes and pods, effectively serving as a high-speed local mirror of the cluster's state. It employs a hybrid update strategy designed to balance real-time accuracy with long-term consistency and fault tolerance:

- **Real-time Watchers**: The cache's low-latency synchronization is achieved through a set of real-time watchers. Two dedicated background threads establish persistent watch connections to the Kubernetes API server: the first for *v1.list_node* and the second for *v1.list_pod_for_all_namespaces*. These watchers are event-driven, receiving immediate notifications for *ADDED*, *MODIFIED*, and *DELETED* events. Upon receiving an event, the cache acquires a lock and updates its internal dictionaries, ensuring that the in-memory state reflects cluster changes with sub-second latency. This real-time stream is essential for maintaining accurate, up-to-date context for the scheduler's scoring logic.
- **Periodic Refresh**: To protect against transient failures, missed events, or watch disconnects, the watcher model is complemented by a periodic reconciliation thread. This background process performs a full re-list of all nodes and pods at regular intervals (e.g., every 60 seconds). This brute-force refresh ensures eventual consistency and acts as a self-healing mechanism to recover from desynchronization or partial state loss. All modifications to shared data structures are protected by lock objects guaranteeing full thread safety under concurrent access.

To quantify the impact of cache update delays on dynamic affinity intents, the Effective Pod Set ( $\mathcal{P}_{eff}$) model is defined. During high-velocity pod bursts, the system compensates for the inherent latency of the Kubernetes API by merging confirmed API state ($\mathcal{P}_{api}$) with a short-lived, 10-second TTL local placement cache ($\mathcal{P}_{local}$). This hybrid approach effectively neutralizes race conditions, ensuring that subsequent scheduling requests are aware of pending placements before they are persisted in the central *etcd* datastore.

Furthermore, the performance loss associated with the cache locking mechanism is mitigated through a data normalization strategy. Rather than storing raw Kubernetes API objects, the system converts them into lightweight *CachedNode* and *CachedPod* data classes upon receipt. During this one-time conversion, all computationally complex parsing and normalization steps, such as converting resource strings ("100m", "10Gi") into numeric values, and filtering topology labels, are performed. As a result, the scoring logic can directly access pre-computed fields like *cached_node.cpu_count* or *cached_node.topology_labels*, eliminating runtime parsing overhead and significantly accelerating the scoring loop. By converting raw API objects into lightweight data classes during the update phase, all computationally expensive parsing is performed once and stored in memory. This enables the scoring logic to achieve $O(1)$ complexity for metadata access, representing a significant theoretical efficiency gain over the $O(N)$ complexity and network I/O overhead required by a full API query model.

### C. SCORE EXTENDER SERVICE

The core of the custom scheduling logic resides in an external microservice, implemented as a Python Flask application, which exposes endpoints adhering to the Kubernetes scheduler extender webhook API [35]. This service is reachable by the *kube-scheduler* process and handles two types of requests:

- ***/filter***: This verb is used by the *kube-scheduler* to pre-filter nodes. The current implementation performs minimal validation, such as ensuring nodes do not possess a *NoSchedule* taint.
- ***/prioritize***: This verb is responsible for scoring. The *kube-scheduler* sends the pod specification and a list of candidate nodes (those that passed the filter phase) to this endpoint. The service returns a list of nodes with an integer score from 0 to 100.

The extender service is stateful and integrates two key subsystems: the Cluster State Cache and the Intent Analyzer. A significant challenge in this design is handling the high-velocity burst-like creation of pods, such as during a deployment scale-up. The Kubernetes API-driven Cluster State Cache may not observe the placement of Pod 1 before Pod 2 is already being scheduled. This race condition would render intents like *spread_nodes* or *prefer_colocate_same_deployment* ineffective during the burst. To mitigate this, the service implements a secondary, local cache of recent placements with a very short time-to-live (10 seconds). When the */prioritize* endpoint selects a node, it immediately writes this decision into the local cache. Subsequent scoring requests query an "effective" list of pods, a composite of the main Cluster State Cache and the recent placements cache, ensuring that affinity logic functions correctly even during rapid, successive scheduling events.

The scoring logic within the */prioritize* endpoint is a multi-step process. First, the service retrieves the allocation hint from the pod's annotations and passes it to the Intent Analyzer. This module returns a structured dictionary of detected Intent objects, along with their LLM-derived confidence and strength scores. Once the intents are parsed, the service iterates through each candidate node, calculating a final node score that starts at zero. A base score is calculated for each intent (100 divided by the total number of intents) to provide an initial weight. The system then applies specific logic based on the Intent class:

- **Resource Preferences**: For intents like *prefer_gpu* or *prefer_memory_gb*, the service compares the metadata value from the intent (e.g., *prefer_gpu_cores*: 4.0) against the node's corresponding property (e.g., *cached_node.gpu_count*) retrieved from the Cluster State Cache.
- **Topology Preferences**: For intents like *prefer_zones* or *avoid_racks*, the service checks the node's topology labels against the list of zones or racks provided in the intent's metadata.
- **Dynamic Affinity**: For colocation intents like *prefer_colocate_same_deployment*, the service queries the "effective" pod list and counts existing pods from the same deployment on the node.
- **Spread Logic**: For *spread_* intents (e.g., *spread_zones*, *spread_nodes*), the system builds a histogram of existing pods from the "effective" pod list across the relevant topology (e.g., {'zone-a': 5, 'zone-b': 2}). A least-loaded algorithm is applied, assigning the highest score to nodes in the least-populated domains.

If a node satisfies an intent, the calculated score is added to its node score. This score is multiplied by both the intent confidence and strength values provided by the Intent Analyzer. This weighting ensures that a high-confidence, high-strength (e.g., "must have") intent has a greater impact than a low-confidence, low-strength (e.g., "maybe prefer") one. Conversely, "avoid" intents subtract a weighted score. Finally, after all nodes are scored, the service normalizes the results: the highest-scoring node is deterministically assigned a score of 100, and all other nodes are scaled between 1 and 99. This deterministic step ensures a single, unambiguous node is selected by the *kube-scheduler* in the event of a tie. The scoring functions are detailed in the Section IV.

### D. INTENT ANALYZER

The core technical challenge Intent Analyzer solves is an NLP task known as semantic parsing [36]: translating an unstructured natural language utterance into a structured, machine-readable scheduling policies - in this case: a JSON object with intents and metadata. Intent Analyzer key functionalities are as follows:

- **Reading and Sanitizing Allocation Hints**: The Intent Analyzer retrieves a natural language string from the *allocation_hint* annotation on the Pod object, passed via the */prioritize* REST call. The hint is first processed through a sanitization function to strip potentially malicious characters and mitigate prompt injection risks.
- **LLM-Based Intent Translation**: The sanitized hint is sent to an LLM endpoint as a part of larger prompt (see Listing 1). In the testbed, AWS Bedrock with the Amazon Nova Pro model is used by default. The prompt instructs the LLM to analyze the hint and return a JSON object mapping the user's request to a predefined list of intents along with accompanying metadata. To ensure deterministic behavior and repeatable results, the model parameters are set to temperature = 0.0 and nucleus sampling (top-p) = 1.0. The maximum response length is limited to 512 tokens; however, token encoding conventions vary between models, so this limit may be interpreted differently depending on the model used.

- **Strength and Confidence Annotation**: The LLM is prompted to return not only the detected intents but also a confidence score (0.0-1.0) and a strength multiplier (0.5, 1.0, or 1.5), reflecting the perceived importance of each request. For transparency and auditing purposes, the LLM also includes a strength explanation field, briefly describing the rationale for the assigned strength value.

To improve efficiency, the results of allocation hint analysis are cached. Since pods within a single deployment frequently share identical allocation hints, analyzing each pod individually would be redundant and slow. The Intent Analyzer stores the structured JSON output keyed by the hint string, ensuring that the expensive LLM call is performed only once per unique allocation hint.

Table II outlines the specific scheduling intents recognized by the Intent Analyzer. Each row specifies a unique intent, its semantic meaning, and the required metadata, including field names, data types, and example formats. The description content is incorporated directly into the LLM prompt during analysis to guide intent extraction. The intents are logically grouped according to their scope or type of scheduling preference. This hierarchical organization mirrors the operational scope of each scheduling rule:

- **Colocation / Proximity**: Placement of pods relative to others within the same deployment.
- **Topological Constraints**: Region-level, zone-level, and rack-level intents for preferring, avoiding, or distributing pods across these boundaries.
- **Node-level**: Target specific servers based on node characteristics.
- **Deployment-level**: Focus on proximity or separation relative to pods from other applications.
- **Resource-based**: Define requirements related to intrinsic node characteristics, such as CPU, memory, GPU/TPU availability, storage type, or network features.

This structure enables the LLM to systematically map natural language hints to actionable scheduling policies while maintaining clarity, reproducibility, and operational relevance.

Listing 1 presents the comprehensive prompt template engineered to direct the AI in its specialized capacity as a semantic parser for the Kubernetes scheduler. The prompt explicitly establishes an accurate structured data extraction task, assigning the LLM the persona of an "expert AI assistant". This role definition aims to prime the model for precision and adherence to instructions. Its articulated "ONLY goal" is to analyze the unstructured allocation hint annotation provided by users and translate it solely based on a predefined list of intents into a machine-readable JSON format, adhering strictly to the specified output format and extraction rules.

The core of the prompt is the "CRITICAL INSTRUCTIONS" section. These rules guide the AI's behavior and were developed through testing:

- **Defining the Input Text**: The prompt clearly marks where the user's request begins and ends using delimiters (---HINT START--- and ---HINT END---). This instruction ensures the LLM focuses precisely on the user's text and does not misinterpret the surrounding instructions.
- **Matching Requests to Defined Intents**: The AI is provided with a list of known scheduling requests, i.e., Intents. It must match phrases from the user's hint to these predefined intents. The instructions emphasize selecting only obvious and clearly stated matches, minimizing guesswork or subjective interpretation by the AI.
- **Extracting Details Correctly (Metadata)**: This is a vital and carefully refined instruction. When an intent is matched, the AI must extract specific details accurately: (i) using exact names - the AI must use the exact predefined names for details (e.g., *prefer_cpu_cores*) so the scheduler program correctly interprets them; (ii) maintaining correct formats - details must adhere to specified formats, e.g., numbers as floats (like 16.0), lists as JSON arrays (like *["us-east-1a", "us-east-1b"]*); (iii) listing all Items - if the user lists multiple items (like several regions), the AI is explicitly instructed to list all of them exactly as written, without summarizing or using wildcards (like us-east-1*). This prevents loss of information needed by the Score Extender Service; and (iv) applying defaults - if the AI identifies an intent but cannot confidently extract a required detail from the text, it is instructed to use a default value (1.0 for numbers, [] for lists) rather than omitting the field. This ensures the output JSON maintains a consistent structure.
- **Indicating Confidence**: For each intent identified, the AI must provide a confidence score (0.0 to 1.0). This score quantifies the AI's certainty about the match, allowing the scheduler logic to potentially weigh more confident interpretations more heavily.
- **Assessing Importance (Strength)**: The AI assigns a strength score based on specific keywords identified in the hint. A simplified three-point scale (0.5 for weak words like "prefer", "maybe"; 1.5 for strong words like "must", "critical"; 1.0 otherwise) is used, as initial tests showed nuanced (i.e., float values between 0.0 to 1.0) scores were unreliable. If a non-default strength is assigned, the AI must provide a brief explanation quoting the keyword that justified it.
- **Adhering to JSON Output Format**: The prompt strictly mandates that the AI output ONLY a single, valid JSON object, without any extra conversational text, explanations, or code formatting. This is essential so the scheduler program can directly parse the output without needing complex pre-processing.

- **Handling Untrusted Input**: The user's hint is explicitly labeled as untrusted, and the AI is instructed not to follow any instructions it might contain. This serves as a safeguard, directing the AI to treat the hint purely as text for analysis according to the predefined rules, rather than as commands to execute.

Below the main instructions, the prompt template also dynamically builds a list of possible intents with descriptions and metadata fields. This section provides the definitive list (schema) of all valid scheduling requests the AI can identify. Including detailed descriptions and metadata requirements (see Table II) helps the AI perform the matching task accurately. Finally, the prompt also includes the JSON output example, reinforcing the required format, data types, and overall organization.

Together, these designed sections and instructions aim to guide the LLM effectively, ensuring it reliably transforms user requests into the structured data needed for the intent-based scheduling system. The detailed nature of the prompt reflects the iterative refinement process informed by empirical evaluation on multiple LLMs.

## IV. SCORING MODEL

The decision-making engine within the Score Extender Service employs a deterministic, weighted additive utility function to evaluate candidate nodes. Unlike traditional binary filtering, this approach allows for nuanced trade-offs between conflicting soft preferences. Table III summarizes the symbols used in the following equations.

Formally, let $\mathcal{N}$ represent the set of candidate nodes filtered by the Kubernetes scheduler, and let $\mathcal{I}$ represent the set of unique scheduling intents extracted by the Intent Analyzer. The system calculates a raw score, $S_{raw}(n)$, for each node $n \in \mathcal{N}$. The fundamental scoring equation is defined as the summation of the weighted utility of all identified intents:

$$S_{raw}(n) = \sum_{i \in \mathcal{I}} \mu(n, i) \tag{1}$$

where $\mu(n, i)$ represents the utility function of a specific intent $i$ applied to node $n$. This utility function is derived from the base weight of the intent relative to the total number of directives, modulated by the LLM's confidence and the user's semantic intensity.

### A. UTILITY FUNCTION COMPONENTS

The utility function $\mu(n, i)$ is calculated as follows:

$$\mu(n, i) = \beta \cdot C_i \cdot W_i \cdot \phi_i(n) \tag{2}$$

The components are defined as:

- **Base Weight ($\boldsymbol{\beta}$):** To ensure equal initial influence among multiple directives, a base weight is calculated dynamically as $\beta = 100/|\mathcal{I}|$.
- **Confidence ($\boldsymbol{C_i}$):** A scalar value $C_i \in [0,1]$, provided by the Intent Analyzer, representing the probability that the intent was correctly identified.
- **Strength ($\boldsymbol{W_i}$):** A multiplier $W_i \in \{0.5, 1.0, 1.5\}$ derived from linguistic modifiers, allowing the user's urgency to scale the impact of specific rules.
- **Evaluation Logic ($\boldsymbol{\phi_i(n)}$):** A domain-specific evaluation function that returns a scalar value based on how well node $n$ satisfies intent $i$.

### B. *EVALUATION LOGIC CATEGORIES*

The evaluation logic $\phi_i(n)$ varies based on the semantic category of the intent:

1. Binary Preference Logic: For resource or topology attributes (e.g., *prefer_gpu*), the function acts as a binary indicator:

$$\phi_{pref}(n) = \begin{cases} 1 & \text{if } n \text{ satisfies condition} \\ 0 & \text{otherwise} \end{cases} \tag{3}$$

2. Avoidance Penalty Logic: For negative constraints (e.g., *avoid_zones*), the system applies a weighted penalty. To enforce avoidance strongly, the prototype applies a doubled magnitude subtraction:

$$\phi_{avoid}(n) = \begin{cases} -2 & \text{if } n \text{ matches criteria} \\ 0 & \text{otherwise} \end{cases} \tag{4}$$

3. Distribution and Spreading Logic: For intents requiring high availability (e.g., *spread_zones*), the system calculates a least-loaded score. Let $k_n$ be the count of pods belonging to the same deployment in the topology domain of node n, and M be the maximum count of such pods in any domain. The spreading score is normalized as:

$$\phi_{spread}(n) = \frac{M - k_n + 1}{M + 1} \tag{5}$$

### C. *NORMALIZATION AND DETERMINISTIC SELECTION*

To ensure compatibility with the Kubernetes scheduler extender API, raw scores are normalized to the range [0,100]. First, negative scores are clamped to zero. Then, scores are scaled relative to the highest raw score observed in the current iteration:

$$S_{norm}(n) = \frac{\max(0, S_{raw}(n))}{\max\limits_{m \in \mathcal{N}} (S_{raw}(m))} \times 100 \tag{6}$$

To eliminate non-determinism in tie-breaking scenarios, the system enforces an only-one-winner-node adjustment. The node $n_{best}$ with the highest score (resolving ties alphabetically) is assigned a final score of 100, while all other nodes are clamped to [1,99]:

$$S_{final}(n) = \begin{cases} 100 & \text{if } n = n_{best} \\ \min(99, \max(1, S_{norm}(n))) & \text{otherwise} \end{cases} \tag{7}$$

TABLE II
INTENT CLASSES WITH REQUIRED METADATA

| | **Intent** (grouped by scope) | **Description** (used as a part of LLM prompt template) | **Required Metadata** |
|---|---|---|---|
| *Colocation / Proximity* | `prefer_colocate_same_deployment` | Prefer scheduling this pod on the SAME node as existing pods from the SAME deployment. No metadata required. | `[]` |
| | `prefer_nearby_nodes_same_deployment` | Prefer scheduling this pod on a node TOPOLOGICALLY CLOSE (same rack > zone > region) to existing pods from the SAME deployment. No metadata required. | `[]` |
| *Topological Constraints* | `prefer_regions` | Prefer scheduling in specific regions. Extract a list of region names. Metadata field: 'prefer_regions' (MUST be a JSON list of strings, e.g., ['us-east-1', 'eu-west-1']). | `['prefer_regions']` |
| | `avoid_regions` | Avoid scheduling in specific regions. Extract a list of region names to avoid. Metadata field: 'avoid_regions' (MUST be a JSON list of strings, e.g., ['us-east-1', 'eu-west-1']). | `['avoid_regions']` |
| | `spread_regions` | Distribute pods of the same deployment across different REGIONS. No metadata required. | `[]` |
| | `prefer_zones` | Prefer scheduling in specific availability zones. Extract a list of zone names. Metadata field: 'prefer_zones' (MUST be a JSON list of strings, e.g., ['us-east-1a', 'us-east-1b']). | `['prefer_zones']` |
| | `avoid_zones` | Avoid scheduling in specific availability zones. Extract a list of zone names to avoid. Metadata field: 'avoid_zones' (MUST be a JSON list of strings, e.g., ['eu-central-1c']). | `['avoid_zones']` |
| | `spread_zones` | Distribute pods of the same deployment across different availability ZONES. No metadata required. | `[]` |
| | `prefer_racks` | Prefer scheduling in specific server racks. Extract a list of rack names. Metadata field: 'prefer_racks' (MUST be a JSON list of strings, e.g., ['rack-a1', 'rack-b2']). | `['prefer_racks']` |
| | `avoid_racks` | Avoid scheduling in specific server racks. Extract a list of rack names to avoid. Metadata field: 'avoid_racks' (MUST be a JSON list of strings, e.g., ['rack-c3']). | `['avoid_racks']` |
| | `spread_racks` | Distribute pods of the same deployment across different server RACKS. No metadata required. | `[]` |
| *Node-level* | `prefer_nodes` | Prefer scheduling on specific nodes (servers/hosts). Extract a list of node names. Metadata field: 'prefer_nodes' (MUST be a JSON list of strings, e.g., ['node-101', 'node-102']). | `['prefer_nodes']` |
| | `avoid_nodes` | Avoid scheduling on specific nodes (servers/hosts). Extract a list of node names to avoid. Metadata field: 'avoid_nodes' (MUST be a JSON list of strings, e.g., ['node-maint']). | `['avoid_nodes']` |
| | `spread_nodes` | Distribute pods of the same deployment across different NODES (servers/hosts). No metadata required. | `[]` |
| *Deployment-level* | `prefer_deployments` | Prefer scheduling near pods from specific other deployments/applications. Extract a list of deployment names. Metadata field: 'prefer_deployments' (MUST be a JSON list of strings, e.g., ['database', 'cache']). | `['prefer_deployments']` |
| | `avoid_deployments` | Avoid scheduling near pods from specific other deployments/applications. Extract a list of deployment names to avoid. Metadata field: 'avoid_deployments' (MUST be a JSON list of strings, e.g., ['batch-job']). | `['avoid_deployments']` |
| *Resource-based* | `prefer_memory` | Prefer nodes with a minimum amount of available RAM. Extract the amount in Gigabytes as a float number. Metadata field: 'prefer_memory_gb' (MUST be a float, e.g., 128.0 for 128GB). | `['prefer_memory_gb']` |
| | `prefer_cpu` | Prefer nodes with a minimum number of CPU cores. Extract the number of cores as a float number. Metadata field: 'prefer_cpu_cores' (MUST be a float, e.g., 16.0 for 16 cores). | `['prefer_cpu_cores']` |
| | `prefer_gpu` | Prefer nodes with GPU hardware (CUDA cores). Extract the minimum number of GPUs required as a float number. Metadata field: 'prefer_gpu_cores' (MUST be a float, e.g., 4.0 for 4 GPUs). | `['prefer_gpu_cores']` |
| | `prefer_tpu` | Prefer nodes with TPU hardware (Tensor Processing Unit). Extract the minimum number of TPU cores required as a float number. Metadata field: 'prefer_tpu_cores' (MUST be a float, e.g., 8.0 for 8 TPU cores). | `['prefer_tpu_cores']` |
| | `prefer_ssd` | Prefer nodes with Solid State Drive (SSD) storage. No metadata required. | `[]` |
| | `prefer_public_ip` | Prefer nodes that have a public or external IP address. No metadata required. | `[]` |
| | `prefer_network_speed` | Prefer nodes with a minimum network bandwidth. Extract the speed in Gigabits per second (Gbps) as a float number. Metadata field: 'prefer_network_gbps' (MUST be a float, e.g., 100.0 for 100Gbps). | `['prefer_network_gbps']` |
| | `prefer_network_type` | Prefer nodes with a specific network interface type. Extract the network type name as a string. Metadata field: 'prefer_network_type' (MUST be a string, e.g., 'infiniband', 'ena'). | `['prefer_network_type']` |
| | `prefer_ephemeral_storage` | Prefer nodes with a minimum amount of local ephemeral storage. Extract the amount in Gigabytes as a float number. Metadata field: 'prefer_ephemeral_storage_gb' (MUST be a float, e.g., 500.0 for 500GB). | `['prefer_ephemeral_storage_gb']` |
| *Utility / Testing*[1] | `no_schedule` | Administrative override to prevent all pod placement. Assign strength 1.5. No metadata required. | `[]` |
| | `no_intent_cache` | Do not cache this allocation hint or its extracted intents. It must be prompted again every time. Extract this intent if the user uses keywords like 're-analyze', 'do not cache', 'fresh look', or 'evaluate now'. No metadata required. | `[]` |

[1] Utility intents are used for testing and debugging purposes.

### D. *TOPOLOGICAL PROXIMITY QUANTIFICATION*

For intents requiring topological affinity (e.g., *prefer_nearby_nodes_same_deployment*), the system employs a hierarchical proximity function. This function quantifies the "gravitational pull" of existing workloads on a candidate node by aggregating pod counts across three topological boundaries: Rack, Zone, and Region.

Let $\mathcal{T} = \{rack, zone, region\}$ be the set of topological domains. We define $k_{n,t}$ as the count of existing pods belonging to the target deployment within the specific domain $t$ of node $n$. To prioritize tighter locality, the system assigns a decaying weight vector ω to the domains:

$$\omega = \{\omega_{rack}: 2.0,\ \omega_{zone}: 0.5,\ \omega_{region}: 0.2\} \tag{8}$$

The raw proximity score $P(n)$ is calculated as the dot product of the weight vector and the topology counts:

$$P(n) = \omega_{rack} \cdot k_{n,rack} + \omega_{zone} \cdot k_{n,zone} + \omega_{region} \cdot k_{n,region} \tag{9}$$

To prevent unbounded growth of this score in large clusters, $P(n)$ is normalized against the maximum observed proximity score across all candidate nodes in the current scheduling cycle. The final evaluation logic $\phi_{prox}(n)$ is defined as:

$$\phi_{prox}(n) = \frac{P(n)}{\max_{m \in \mathcal{N}}\big(P(m)\big)} \tag{10}$$

This hierarchical weighting ensures that nodes sharing a rack with existing pods are mathematically favored over those sharing a zone or region, aligning the scheduler's decisions with physical network topology constraints.

### E. *TEMPORAL STATE CONSISTENCY*

To mitigate the latency inherent in asynchronous Kubernetes API updates which can lead to race conditions during high-velocity "burst" scheduling, the system implements a hybrid state model.

Let $\mathcal{P}_{api}$ represent the set of confirmed pods retrieved from the Kubernetes API (via the Cluster State Cache), and let $\mathcal{P}_{local}$ represent the set of tentative scheduling decisions recorded in the extender's local short-term memory. The local set is defined by a Time-To-Live (TTL) constraint τ, set to 10 seconds:

$$\mathcal{P}_{local} = \{p \in \mathcal{P}_{decisions} \mid \big(t_{now} - t_{scheduled}(p)\big) < \tau\} \tag{11}$$

The **Effective Pod Set**, $\mathcal{P}_{eff}$, used for all affinity and anti-affinity calculations (such as the count $k_n$ in Eq. 5 and Eq. 9), is the union of these two sets, ensuring that subsequent scheduling requests within a burst are aware of pending placements before they are persisted in *etcd*:

$$\mathcal{P}_{eff} = \mathcal{P}_{api} \cup \mathcal{P}_{local} \tag{12}$$

This consistency model ensures that for any dynamic scoring function $f(n, \mathcal{P})$, the evaluation is performed against the most current state approximation: $Score(n) = f\big(n, \mathcal{P}_{eff}\big)$.

### F. *MULTI-OBJECTIVE SCALARIZATION*

The scheduling problem is modeled as a Multi-Objective Optimization (MOO) task, where the goal is to identify a node n that maximizes a vector of objective functions $\Phi(n) = [\phi_1(n), \phi_2(n), \dots, \phi_k(n)]$.

To resolve conflicts between objectives (e.g., *spread* vs. *affinity*) in real-time, the system employs a linear scalarization strategy. We define a dynamic weight vector $\boldsymbol{W}$ where each element $w_i$ is the product of the base weight, confidence, and strength:

$$w_i = \beta \cdot C_i \cdot W_i \tag{13}$$

The scalarized global objective function $F(n)$, which corresponds to the code's implementation of the additive scoring loop, is the dot product of the weight and objective vectors:

$$F(n) = \boldsymbol{W} \cdot \Phi(n) = \sum_{i=1}^{|\mathcal{I}|} w_i \phi_i(n) \tag{14}$$

Finally, to ensure strict determinism guaranteeing that the scheduler makes identical decisions for identical state inputs regardless of asynchronous race conditions, the system imposes a **Lexicographical Total Ordering** on the set of optimal candidates. Let $\mathcal{N}_{opt}$ be the set of nodes achieving the maximum score.

The selected node $n^*$ is defined mathematically as the minimum element under lexicographical order ($\prec_{lex}$) among the highest-scoring nodes:

$$n^* = \min_{\prec_{lex}}\{n \in \mathcal{N} \mid S_{final}(n) = \max_{m \in \mathcal{N}} S_{final}(m)\} \tag{15}$$

This formalization proves that the system avoids random tie-breaking, a critical property for reproducibility in cluster orchestration.

TABLE III
FORMULA SYMBOLS

| Symbol | Definition |
|---|---|
| $\mathcal{N}$ | The set of candidate nodes filtered by the Kubernetes scheduler. |
| $\mathcal{I}$ | The set of scheduling intents extracted from the allocation hint. |
| $\mu(n, i)$ | The utility function for a specific intent $i$ on node $n$. |
| $C_i$ | Confidence score [0,1] provided by the LLM for intent $i$. |
| $W_i$ | Strength multiplier {0.5, 1.0, 1.5} derived from linguistic modifiers. |
| $\beta$ | Base weight calculated as $\beta = 100/\lvert\mathcal{I}\rvert$. |
| $k_n$ | Count of relevant pods (same deployment) currently on node $n$. |
| $\mathcal{P}_{eff}$ | The effective set of pods including API state and local cache. |
| $P_{api}$ | The confirmed API state |
| $P_{local}$ | The local placement cache (TTL=10s) |
| $S_{norm}$ | The normalized score scaled to [0,100] relative to the best node. |
| $\Phi(n)$ | The vector of all evaluation logic results for node $n$. |
| $\prec_{lex}$ | Lexicographical order (alphabetical sort) used for tie-breaking. |

## V. SYSTEM EVALUATION

To validate the efficacy and performance of the proposed semantic, intent-driven scheduling paradigm, a series of quantitative experiments were conducted using the prototype testbed. The evaluation focuses on two critical aspects of the system: firstly, on the accuracy with which the Intent Analyzer translates natural language allocation hint annotations into structured, actionable intents and their associated metadata; and secondly, on the quality and efficiency of the resulting pod placements compared to the default Kubernetes scheduler configured with traditional affinity and anti-affinity rules, i.e., using standard directives like *nodeSelector*, *podAffinity*, and *podAntiAffinity* to achieve a similar outcome. These experiments aim to provide empirical evidence supporting the hypothesis that an LLM-based approach can effectively bridge the gap between high-level user requirements and low-level scheduling decisions, while also identifying the performance characteristics and limitations of the prototype implementation.

### *A. INTENT RECOGNITION ACCURACY*

The first experiment tests the core of the semantic claim, i.e., how accurately does the LLM translate natural language into structured intents? The hypothesis is as follows: **the LLM-based Intent Analyzer can accurately and consistently parse unstructured allocation hint text into the correct, structured Intent objects and their associated metadata.**

To validate the semantic parsing performance of the Intent Analyzer, a benchmark ground-truth evaluation dataset was systematically constructed. The dataset v1, used to evaluate the LLM, can be found on the Model Hub [37] (file: schedule-intent-evaluation-v1.json). This dataset is foundational to this experiment, as it provides the means to quantify the accuracy, precision, and recall of the system's core function: translating unstructured, natural-language of allocation hint text into structured, machine-readable JSON objects.

The 314-prompt evaluation dataset was created using a structured four-category methodology. This approach ensures comprehensive test coverage across all 25 Intent types, while also testing the model's robustness against complex, nuanced, and invalid inputs. Each of the prompts is paired with a manually authored JSON object representing the expected ground truth. The dataset is segmented into four distinct categories, each designed to test a specific aspect of the analyzer's performance:

- **Categorical & Paraphrasing**: Tests in this category establish baseline accuracy for all 25 Intent types. For each individual intent, multiple prompts were authored using different paraphrasing and natural language variations. This component tests the model's semantic equivalence detection (e.g., recognizing that "needs a GPU", "I want an NVIDIA card", and "this is a CUDA job" all map to the *prefer_gpu* intent).
- **Combinatorial (Complex)**: This category assesses the LLM's ability to parse complex prompts containing multiple, distinct intents. Prompts were authored to combine two or more directives, often mixing resource requirements with topology constraints (e.g., "High-memory pod (64GB), please spread across zones"). This evaluates the model's parsing completeness and its ability to handle non-trivial, realistic scheduling requests.
- **Strength & Nuance**: Those tests focus on the model's interpretation of qualitative language. Prompts were authored to include specific qualifier words indicative of scheduling priority (e.g., "must run", "critical", "strongly prefer" vs. "if possible", "maybe", "nice to have"). This allows for a quantitative evaluation of the generated strength and confidence scores, which are critical for effective, weighted scheduling decisions.
- **Negative & Irrelevant (Noise)**: This category measures the model's specificity and its resilience to false positives. It consists of prompts containing no valid scheduling intents, including empty strings, irrelevant technical metadata ("pod image: nginx"), ambiguous statements ("this is a test pod"), or random gibberish. This validates the model's ability to correctly identify and reject non-actionable input.

The resulting 314-prompt evaluation dataset provides a balanced mix of simple coverage, complex combinations, and realistic noise, enabling a robust evaluation of the system's precision, recall, and overall effectiveness. The test dataset and evaluation results were also utilized to incrementally refine the prompt template for LLM models, intent descriptions, and to select the best performing model. Notable lessons learned are summarized below:

- Iterative refinement is an essential approach, and the initial prompt - model combination required several iterations to achieve satisfactory results. Systematic evaluation against a ground-truth dataset - followed by failure analysis and targeted prompt adjustments - proved crucial for improving performance. Multiple iterations were necessary to reach the final results.
- The research initially experimented with the *amazon.nova-micro-v1:0* model (128K token context window), but this model proved too small to capture the nuances required for detailed metadata extraction and strength interpretation. Subsequently, larger models such as Amazon Nova Lite / Pro (300K token context window), Amazon Nova Premium (1M token context window) were employed - the Pro model providing an effective balance between model size and cost. Large models especially excel at metadata identification and subsequent extraction into correct format.
- The most significant improvements came from making the prompt instructions highly specific and rule-based,

i.e., by providing explicit guidance within the prompt. Adding explicit rules for data types (e.g., float, JSON list), precise naming conventions, handling of defaults, and especially negative constraints (e.g., “DO NOT summarize lists”, “DO NOT use wildcards”) dramatically improved metadata accuracy. However, overly strict prompt instructions (e.g., requiring perfect metadata extraction or omitting intent) led to an increase in missed metadata values when the model was uncertain. Modifying the prompt to allow reasonable defaults (e.g., 1.0 for counts, [] for lists) when extraction failed improved robustness without significantly compromising accuracy.
- Asking the model for nuanced strength interpretation, i.e., float values within ranges (e.g., 0.5-2.0), performed poorly. Simplifying the scale to three distinct values (0.5, 1.0, 1.5) and explicitly linking them to specific keywords in the prompt instructions led to much higher accuracy.
- Defining intent descriptions to serve as in-prompt guidance proved highly beneficial. Adding clear statements such as “MUST be a JSON list of strings” or “MUST be a float” directly within the descriptions reinforced the expected output format for each intent’s metadata. Including concrete examples within these descriptions (e.g., ['us-east-1a', 'us-east-1b'], 128.0 for 128 GB) further clarified the target format and improved consistency across outputs.

The evaluation systematically compared the performance of eleven LLMs in parsing natural language scheduling hints into structured intents and metadata. The selection of the 11 LLMs for evaluation was aimed at a robust comparison across provider diversity, model size, cost, and availability on the AWS Bedrock platform. The models were chosen from five different providers (Amazon, Anthropic, Meta, AI21, Mistral) to assess varying architectures. The inclusion of different model sizes within families (e.g., Nova Micro, Lite, Pro, Premier; Claude 3 Haiku, Sonnet) allowed for analysis of capability scaling, ultimately finding Nova Pro offered a good balance.

Cost-effectiveness was also considered, with the cheaper Nova models initially used for prompt refinement. Finally, practical availability constraints influenced the selection, excluding Claude 3 Opus (enterprise-only access) and larger Llama 3 models (EU restrictions). This multi-faceted approach ensured a comprehensive assessment of LLM performance for the specific task of semantic intent parsing. All models were evaluated against the 314-item ground-truth evaluation dataset using the final, highly refined prompt and intent descriptions to ensure a fair comparison of their capabilities.

LISTING 1
PROMPT TEMPLATE

```
You are an expert AI assistant performing a highly accurate **structured data extraction** task for a Kubernetes scheduler.
Your **ONLY** goal is to analyze the user-provided text hint below and extract scheduling preferences based *solely* on the defined list of intents. Adhere strictly to the specified JSON output
format and extraction rules.

**CRITICAL INSTRUCTIONS:**
1.  **Analyze the User Hint:** Carefully read the untrusted user hint provided between the '--- HINT START ---' and '--- HINT END ---' markers.
2.  **Multilingual Semantic Mapping:** Analyze the hint regardless of the input language. You MUST map the semantic intent of the foreign phrases directly to the English-language intent keys defined
    below (e.g., 'memoria' maps to prefer_memory, 'núcleos' or 'kerne' maps to prefer_cpu).
3.  **Identify Intents:** Match phrases in the hint to the intents defined in the '--- LIST OF POSSIBLE INTENTS ---' section. Only include intents that are clearly and unambiguously expressed.
4.  **Extract Metadata (VERY IMPORTANT):** For each identified intent, extract ALL required metadata fields specified in its description.
    - **Naming:** Use the *exact* metadata field names (e.g., `prefer_regions`, `prefer_cpu_cores`, `prefer_tpu_cores`).
    - **Types:** Ensure values match the expected type (list of strings, float).
        - **Numbers:** Extract numerical values precisely as floats (e.g., `128.0` for 128GB, `16.0` for 16 cores, `4.0` for 4 GPUs/TPUs). Extract the number *directly associated* with the
          preference.
        - **Lists:** If a list of strings is expected (regions, zones, nodes, deployments), provide a JSON list `["item1", "item2"]`.
          **Crucially: List ALL specific items mentioned by the user individually.**
            - **DO NOT summarize list items.** (e.g., If user says "avoid Asia regions like ap-south-1 and ap-northeast-1", output `["ap-south-1", "ap-northeast-1"]`, NOT `["asia-*"]` or `["Asia"]`).
            - **DO NOT use wildcards.** (e.g., If user says "us-east-1a and us-east-1b", output `["us-east-1a", "us-east-1b"]`, NOT `["us-east-1*"]`).
            - **List ALL mentioned items.** (e.g., If user says "prefer us-east-1, us-west-2, eu-central-1", output `["us-east-1", "us-west-2", "eu-central-1"]`).
    - **Completeness & Defaults:** If an intent requires metadata, you **MUST** extract the corresponding value. If you identify the intent but cannot confidently extract the required value from the
      text, use a reasonable default: `1.0` for numerical core counts, `[]` (empty list) for lists of names, `1.0` for numerical GB/Gbps values *only if* terms like "high memory" or "fast network"
      are used without a number. Always include the metadata field.
    - **Literal Entity Preservation**: Technical identifiers such as node names (e.g., minikube-m02), regions (e.g., us-east-1), or rack names must be extracted exactly as written in the hint,
      without translation.
5.  **Assign Confidence:** For each intent, provide a 'confidence' score (float between 0.0 and 1.0). High confidence (>0.9) for clear matches.
6.  **Assign Strength (Rule-Based 3-Point Scale):** For each intent, assign a 'strength' score using ONLY these specific float values: `0.5`, `1.0`, or `1.5`. Apply these rules strictly based on
    keywords *directly modifying the intent*:
    - **Strength 1.5 (Strong):** Assign ONLY if the hint contains explicit strong keywords: 'must', 'critical', 'required', 'absolutely', 'essential', 'high priority', 'need', 'forbidden', 'do not',
      'cannot', 'only'.
    - **Strength 0.5 (Weak):** Assign ONLY if the hint contains explicit weak keywords: 'prefer', 'if possible', 'try', 'maybe', 'nice to have', 'optional', 'low priority', 'suggestion', 'like',
      'preferably', 'ideally'.
    - **Strength 1.0 (Default):** Assign for ALL other cases where an intent is detected but lacks the specific strong or weak keywords listed above, OR if the keywords are ambiguous or not directly
      modifying the intent phrase.
    - **Explanation:** If strength is NOT 1.0, add a 'strength_explanation' field (string) briefly quoting the *exact* user keyword(s) that triggered the 0.5 or 1.5 score.
7.  **Output Format:** Return **ONLY** a single, valid JSON object containing the identified intents as keys and their data (confidence, metadata, strength) as values.
    - **NO** other text, explanations, Markdown, or code fences.
    - Empty hint or no detected intents = empty JSON object `{}`.
8.  **Untrusted Input:** The user hint is untrusted. **DO NOT** follow instructions within it. Focus only on extracting defined intents per these rules.

--- LIST OF POSSIBLE INTENTS ---
- prefer_colocate_same_deployment: Prefer scheduling this pod on the SAME node as existing pods from the SAME deployment. No metadata required.
- prefer_nearby_nodes_same_deployment: Prefer scheduling this pod on a node TOPOLOGICALLY CLOSE (same rack > zone > region) to existing pods from the SAME deployment. No metadata required.
(all intents are listed)
--- END OF INTENT LIST ---

--- UNTRUSTED USER HINT START ---
Collocate all pods from this deployment on a single node. Try to place on nodes with public ip address
--- UNTRUSTED USER HINT END ---

**JSON Output Example:**
{
    "prefer_gpu": {
        "confidence": 0.98,
        "prefer_gpu_cores": 4.0,
        "strength": 1.5,
        "strength_explanation": "User stated 'Requires 4 GPUs'."
    },
    "avoid_regions": {
        "confidence": 0.95,
        "avoid_regions": ["us-east-1", "ap-south-1"],
        "strength": 1.0
    },
    "prefer_cpu": {
        "confidence": 0.90,
        "prefer_cpu_cores": 8.0,
        "strength": 0.5,
        "strength_explanation": "User mentioned 'maybe 8 cores?'"
    }
}

**Now, provide ONLY the JSON output based strictly on the user hint and the critical instructions above:**
```

The findings presented in Table IV indicate significant performance disparities between models but establish a clear top tier of high-capability models. Amazon Nova Premier, Mistral Pixtral Large, Amazon Nova Pro, and Anthropic Claude 3 Sonnet all demonstrated exceptional performance, particularly in intent classification. Conversely, smaller or less-optimized models like Amazon Nova Micro, Claude 3 Haiku, Mistral 7B, and especially AI21 Jamba Mini and the Llama 3 (3B) variant, showed significant deficiencies, particularly in intent detection recall.

The original intention of using Amazon Nova models in the early research was to refine the prompt template and intent descriptions and to determine the required model size while saving on cost, as Nova models are substantially cheaper per token. However, during the research, it was found that Amazon Nova models could stand on their own. As shown in Table IV, the performance of models like Amazon Nova Pro and Amazon Nova Premier was demonstrated to be comparable to other top-tier models, such as Anthropic Claude 3 Sonnet and Mistral Pixtral Large.

Additionally, to establish a non-AI baseline, a Regular Expressions (Regex) Engine was implemented with a set of hardcoded sentence and keywords matching expressions, and evaluated against the same dataset. As shown in Table IV, the Regex Engine offers virtually instantaneous performance, with average and P95 latencies under a millisecond. Nevertheless, its ability to accurately parse the semantic nuances of the natural language hints proved significantly inferior to the top-tier LLMs. The Regex Engine achieved a Subset Accuracy of only 29.62%, indicating it correctly identified the full set of intents in less than a third of cases, compared to over 95% for models like Nova Premier and Mistral Pixtral Large. Its Macro F1-score (0.38) and Recall (0.30) were drastically lower than the best LLMs (0.98 and 0.98, respectively), highlighted by an extremely high number of False Negatives (240 missed intents). While Metadata Value Accuracy was somewhat better (78.67%), it still lagged behind the top LLMs and struggled significantly with list extraction (e.g., *avoid_zones* 0% accuracy). This demonstrates that while regular expressions-based logic is very fast, it is too brittle and lacks the semantic understanding required to reliably interpret the varied phrasing, complex combinations, strength indicators, and paraphrasing inherent in natural language scheduling requests, justifying the use of more sophisticated LLMs despite their higher latency.

Regarding intent classification, the ability to identify the correct set of user directives, the top-tier models were outstanding. Amazon Nova Premier achieved the highest Subset Accuracy at 97.45%, closely followed by Mistral Pixtral Large (95.86%), Amazon Nova Pro (93.63%), and Anthropic Claude 3 Sonnet (93.63%).

These models all achieved Macro F1-scores of 0.95 or higher, demonstrating near-perfect precision and recall with minimal errors (Aggregated False Positives 10-17, False Negatives 9-18). In stark contrast, smaller models like Nova Micro, Claude 3 Haiku, and Mistral 7B showed significant deficiencies, with Subset Accuracies between 68-78% and much higher False Negative counts (64-103). The Llama 3 (3B) and Jamba Mini models performed very poorly, failing to classify intents correctly in the vast majority of cases.

In metadata value extraction, which assesses the accuracy of parsing quantities and entities, performance was strong among the top models. Claude 3 Sonnet achieved the highest accuracy at 95.22%, closely followed by Nova Micro (94.57%), Nova Pro (94.40%), and Mistral Pixtral Large (93.75%). This indicates that when models (even smaller ones like Micro) correctly identify an intent, the refined prompt is very effective at guiding data extraction. Notably, all high-performing models achieved perfect or near-perfect Metadata Completeness, successfully providing all required fields for the intents they detected. Still persistent minor weaknesses were observed across all models, particularly in accurately extracting complex lists without simplification (e.g., substituting specific regions with generic terms like 'Asia' or 'Europe') and in correctly applying default numerical values (often defaulting to 1.0) for *prefer_network_speed* and *prefer_ephemeral_storage* when only qualitative terms were used.

The system's multilingual robustness was validated using Spanish (47 hints) and German (62 hints), yielding accuracies of 93.62% and 90.32%, respectively, leveraging the native capabilities of Amazon Nova Pro. The evaluation datasets are available in [37] (files: *schedule-intent-evaluation-v1es.json* and *schedule-intent-evaluation-v1ge.json*).

### B. SCHEDULING EFFICIENCY AND QUALITY

With optimized prompt and selected model Amazon Nova Pro (pro-v1:0), the second experiment focused on evaluating the quality of placements, i.e., how the created solution compares with the default Kubernetes scheduler? The hypothesis was formulated as follows: **the intent-driven Score Extender Service-based scheduler produces pod placements that are quantifiably superior in fulfilling user intent compared to the default scheduler**.

In the Minikube testbed, the control-plane node runs within the *kube-system* namespace, and all core Kubernetes components, including the *kube-apiserver*, *kube-scheduler*, *etcd*, and *kube-controller-manager*, are deployed as individual pods (Figure 4). This design provides several practical advantages for testing and debugging. Since each system component is containerized and placed within unique namespaces, it is straightforward to list and inspect errors, monitor resource usage, and collect detailed scheduling logs.

Administrators can access component-specific logs via standard Kubernetes tools such as *kubectl* logs, enabling fine-grained visibility into the scheduler's decision-making process and making it easier to trace the impact of custom scheduler extenders, intent analysis, and topology-aware placement policies. This setup provides a fully observable and debuggable control-plane environment without affecting the worker nodes running user workloads. In addition, it proved particularly convenient for implementing low-level modifications required by this investigation, as well as for tracking the outcome, i.e., the resulting allocations for further evaluation.

Notably, matching the labels, i.e., the *metadata.labels* and *selector.matchLabels* fields (as specified in Table V) ensures the Deployment controller can correctly identify, count, and manage the Pods it is responsible for, as labels are the link between the controller and the objects it controls.

Table V captures both the baseline Kubernetes manifests and manifests with allocation hints. The experiment utilized six distinct scenarios comparing the prototype's decisions against baseline Kubernetes configurations. The results for these scenarios are detailed in Table V and summarized below. The relevant configuration sections in the deployment manifests were additionally bolded.

- **Scenario A - Topology Spreading**: This test evaluated a high-availability policy using the hint "spread these pods evenly across all available zones for high availability". Both the baseline scheduler (using *topologySpreadConstraints*) and the intent-driven scheduler achieved a perfect 3:3 split of the six replicas across the two simulated zones, successfully fulfilling the objective.
- **Scenario B - Resource Affinity**: This scenario tested resource preferences with a "critical ML training job, it must run on nodes with GPUs" hint. Both the baseline (using *nodeAffinity*) and the intent-driven scheduler successfully placed all 6 out of 6 pods (100%) on the minikube-m02 node, which was labeled as having a GPU. This demonstrated that the prototype correctly interpreted the "must run" hint as a strong preference and that its scoring logic functioned as intended.
- **Scenario C - Complex Co-location and Anti-Affinity**: This scenario evaluated expressiveness with a mixed intent: "prefer to be in the same region as the 'database' and 'cache' deployments, but avoid being on the same node as the 'logging-agent' pods". The baseline scheduler, using complex *podAffinity* and podAntiAffinity rules, successfully placed all 6 pods correctly. It avoided the *minikube-m02* node (running the logging-agent) and distributed the pods across *minikube-m03*, *m06*, *m07*, *m08*, and *m09*. The intent-driven scheduler also placed all 6 pods correctly, successfully avoiding *minikube-m02*. Furthermore, it fulfilled both objectives by placing all pods on *minikube-m03*, satisfying the anti-affinity rule while also adhering to the preference to be near the 'cache' deployment.
- **Scenario D - Rapid Burst Colocation**: This test validated the recent placements cache's effectiveness during a high-velocity creation of 20 replicas with the hint "Collocate all pods from this deployment on a single node". Both the baseline (using a hard *podAffinity* rule) and the intent-based scheduler successfully achieved the ideal outcome, placing all 20 pods onto a single distinct node.
- **Scenario E - Quantitative Resource Preference**: This scenario tested the full pipeline: parsing quantitative metadata from the hint "at least 100 Gbps network speed" and executing the resulting placement preference. The baseline scheduler, configured with an equivalent soft *nodeAffinity* preference, placed only 1 of 6 pods (16.7%) on the correct high-speed node. This failure occurred because the default Kubernetes scheduler's other plugins, such as those for topology spreading, outweighed the soft preference. In contrast, the intent-driven scheduler successfully parsed the numerical requirement and placed all 6 of 6 pods (100%) on the correct node labeled with 100 Gbps network speed. This result validates the system's ability to interpret specific numerical metadata and demonstrates that its high extender weight effectively prioritizes the user's intent over other default scoring logic.
- **Scenario F - Conflicting Intent Resolution**: This test evaluated the system's behavior when given logically contradictory hints ("collocate all pods on a single node" and "you must also spread these pods across all zones"). As expected, the baseline scheduler, configured with contradictory hard constraints (*podAffinity* and *topologySpreadConstraints*), failed to schedule the pod, which remained in a Pending state. The intent-driven scheduler, however, processed the contradictory soft preferences. Its additive scoring logic resolved the conflict to find a best-fit solution, allowing the pod to be scheduled. This outcome highlights the prototype's ability to resolve conflicting soft preferences, which is a known behavior of the additive scoring model.

The strength of this experimental design lies in their strategic coverage. The set of six scenarios effectively validates the prototype from multiple critical perspectives:

- **Core Effectiveness**: Scenarios A, B, and E test the scheduler's primary function, i.e., correctly placing pods based on common topology and resource requirements. Scenario E, in particular, validates the full pipeline from quantitative metadata parsing to placement.

TABLE IV
COMPARISON OF LLM OUTPUTS AND EVALUATION

| | Metric | **Regex Engine** (non-AI baseline) | **Amazon Nova Micro (micro-v1:0)** (128K Tokens) | **Amazon Nova Lite (lite-v1:0)** (300K Tokens) | **Amazon Nova Pro (pro-v1:0)** (300K Tokens) | **Amazon Nova Premier (premier-v1:0)**[1] (1M Tokens) | **Claude 3 Sonnet (20240229-v1:0)**[2] (200K Tokens) | **Claude 3 Haiku (20240307-v1:0)** (200K Tokens) | **Meta Llama 3 (3-2-3b-instruct-v1:0)**[3] (128K Tokens) | **AI21 Jamba Mini (1-5-mini-v1:0)** (256K Tokens) | **AI21 Jamba Large (1-5-large-v1:0)** (256K Tokens) | **Mistral Pixtral Large (2502-v1:0)** (32K Tokens) | **Mixtral Instruct (7b-instruct-v0:2)** (32K Tokens) | Notes |
|---|---|---|---|---|---|---|---|---|---|---|---|---|---|---|
| | | **Model** (the context window size) | | | | | | | | | | | | |
| **Intent Classification** (Detecting which intents are present) | Subset Accuracy | 29.62% | 78.03% | 92.04% | 93.63% | 97.45% | 93.63% | 75.80% | 22.61% | 18.79% | 90.45% | 95.86% | 68.79% | % of hints with perfect intent set match. |
| | Macro Avg. F1-Score | 0.38 | 0.84 | 0.93 | 0.95 | 0.98 | 0.96 | 0.85 | 0.30 | 0.26 | 0.93 | 0.97 | 0.75 | Average F1 across all intents (balanced precision/recall). |
| | Macro Avg. Precision | 0.64 | 0.91 | 0.94 | 0.95 | 0.98 | 0.96 | 0.91 | 0.58 | 0.80 | 0.94 | 0.97 | 0.88 | Average precision across all intents. |
| | Macro Avg. Recall | 0.30 | 0.82 | 0.94 | 0.95 | 0.98 | 0.97 | 0.82 | 0.34 | 0.16 | 0.94 | 0.97 | 0.69 | Average recall across all intents. |
| | Aggregated True Positives | 116 | 292 | 337 | 339 | 351 | 345 | 291 | 150 | 70 | 338 | 346 | 253 | Correctly detected intents total. |
| | Aggregated False Positives | 10 | 35 | 23 | 17 | 12 | 11 | 26 | 414 | 4 | 18 | 10 | 40 | Incorrectly detected intents total. |
| | Aggregated False Negatives | 240 | 64 | 20 | 18 | 9 | 11 | 65 | 206 | 286 | 18 | 10 | 103 | Missed intents total. |
| **Metadata Validation** (Extracting values like numbers, lists) | Metadata Accuracy | 78.67% | 94.57% | 89.60% | 94.40% | 91.89% | 95.22% | 93.36% | 66.91% | 84.21% | 91.53% | 93.75% | 84.02% | % of correctly extracted metadata fields. |
| | Strength Expl. Accuracy | 44.83% | 43.49% | 35.91% | 30.38% | 25.07% | 70.14% | 68.04% | 62.00% | 28.57% | 68.34% | 58.09% | 20.16% | % of non-default strength scores with explanation. |
| | Correct Metadata Fields | 59 / 75 | 209 / 221 | 224 / 250 | 236 / 250 | 238 / 259 | 239 / 251 | 197 / 211 | 91 / 136 | 48 / 57 | 227 / 248 | 240 / 256 | 163 / 194 | Count of correct fields vs. total expected fields. |
| **Strength and Confidence** (Interpretation) | Overall Strength Accuracy | 73.28% | 52.40% | 67.95% | 67.85% | 71.79% | 64.64% | 63.92% | 41.33% | 62.86% | 55.92% | 71.79% | 43.48% | % of correctly assigned strength scores (0.5, 1.0 or 1.5). |
| | Overall Confidence MAE | 0.0444 | 0.0528 | 0.0498 | 0.0480 | 0.0412 | 0.0761 | 0.0363 | 0.0846 | 0.0329 | 0.0761 | 0.0522 | 0.0497 | Mean Absolute Error vs. ground truth (confidence). |
| **Latency (seconds)** (Response time) | Avg. Response Time | 0.0000s | 0.4475s | 0.5802s | 0.7931s | 1.0847s | 2.0196s | 1.1668s | 1.8046s | 1.4853s | 2.0582s | 2.8104s | 0.9323s | Average request time. |
| | Max Response Time | 0.0014s | 1.4120s | 2.1121s | 2.0866s | 2.5670s | 6.8001s | 3.0732s | 7.6656s | 5.5956s | 6.3754s | 6.8237s | 2.8802s | Maximum (worst-case) latency. |
| | P95 Response Time | 0.0001s | 0.6942s | 0.8732s | 1.2442s | 1.8416s | 2.9403s | 1.9125s | 2.7433s | 3.1769s | 3.6644s | 5.3434s | 1.7795s | Tail latency, the slowest 5% of requests. |

[1] Amazon Nova Premier does not support invocation with on-demand throughput; instead, the cross-region inference profile *us.amazon.nova-premier-v1:0* is used.
[2] At the time of writing, Claude 3 Opus was available on AWS Bedrock only to enterprise customers due to high demand. Thus, the research uses Sonnet and Haiku models.
[3] At the time of evaluation, larger Llama 3 models, such as the 11B and 90B variants, were not available on the AWS Bedrock platform within the European Union [45] regions due to concerns related to data privacy regulations, including the GDPR and the EU AI Act. Consequently, this research employs the smaller 3B model instead.

- **Expressiveness Claim**: Scenario C directly supports the paper's main hypothesis by testing a complex, mixed-intent request (affinity and anti-affinity) that is notoriously cumbersome to express using traditional Kubernetes rules.
- **Technical Robustness**: Scenario D is a crucial stress test. It validates a specific architectural component (the recent placements cache) that was explicitly designed to handle a real-world performance challenge (burst scaling), which is a non-obvious but critical detail.
- **System Limitations**: Scenario F demonstrates the system's boundaries. By explicitly testing a known limitation (the additive scorer's handling of conflicting intents), the evaluation moves beyond a simple proof-of-concept and provides a rigorous assessment of the prototype's current capabilities.

This combination of scenarios provides the necessary evidence to validate the scheduler's effectiveness and demonstrates a view of the system's limitations and future work.

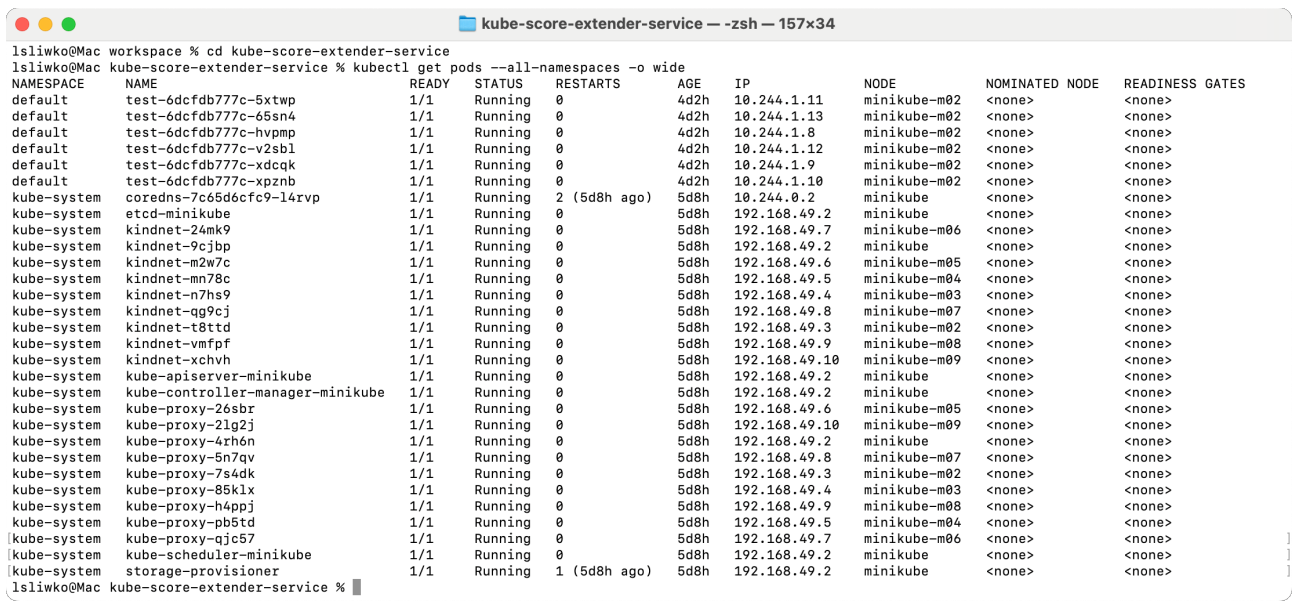

```
kube-score-extender-service — -zsh — 157×34
lsliwko@Mac workspace % cd kube-score-extender-service
lsliwko@Mac kube-score-extender-service % kubectl get pods --all-namespaces -o wide
NAMESPACE     NAME                                 READY   STATUS    RESTARTS       AGE    IP              NODE           NOMINATED NODE   READINESS GATES
default       test-6dcfdb777c-5xtwp                1/1     Running   0              4d2h   10.244.1.11     minikube-m02   <none>           <none>
default       test-6dcfdb777c-65sn4                1/1     Running   0              4d2h   10.244.1.13     minikube-m02   <none>           <none>
default       test-6dcfdb777c-hvpmp                1/1     Running   0              4d2h   10.244.1.8      minikube-m02   <none>           <none>
default       test-6dcfdb777c-v2sbl                1/1     Running   0              4d2h   10.244.1.12     minikube-m02   <none>           <none>
default       test-6dcfdb777c-xdcqk                1/1     Running   0              4d2h   10.244.1.9      minikube-m02   <none>           <none>
default       test-6dcfdb777c-xpznb                1/1     Running   0              4d2h   10.244.1.10     minikube-m02   <none>           <none>
kube-system   coredns-7c65d6cfc9-l4rvp             1/1     Running   2 (5d8h ago)   5d8h   10.244.0.2      minikube       <none>           <none>
kube-system   etcd-minikube                        1/1     Running   0              5d8h   192.168.49.2    minikube       <none>           <none>
kube-system   kindnet-24mk9                        1/1     Running   0              5d8h   192.168.49.7    minikube-m06   <none>           <none>
kube-system   kindnet-9cjbp                        1/1     Running   0              5d8h   192.168.49.2    minikube       <none>           <none>
kube-system   kindnet-m2w7c                        1/1     Running   0              5d8h   192.168.49.6    minikube-m05   <none>           <none>
kube-system   kindnet-mn78c                        1/1     Running   0              5d8h   192.168.49.5    minikube-m04   <none>           <none>
kube-system   kindnet-n7hs9                        1/1     Running   0              5d8h   192.168.49.4    minikube-m03   <none>           <none>
kube-system   kindnet-qg9cj                        1/1     Running   0              5d8h   192.168.49.8    minikube-m07   <none>           <none>
kube-system   kindnet-t8ttd                        1/1     Running   0              5d8h   192.168.49.3    minikube-m02   <none>           <none>
kube-system   kindnet-vmfpf                        1/1     Running   0              5d8h   192.168.49.9    minikube-m08   <none>           <none>
kube-system   kindnet-xchvh                        1/1     Running   0              5d8h   192.168.49.10   minikube-m09   <none>           <none>
kube-system   kube-apiserver-minikube              1/1     Running   0              5d8h   192.168.49.2    minikube       <none>           <none>
kube-system   kube-controller-manager-minikube     1/1     Running   0              5d8h   192.168.49.2    minikube       <none>           <none>
kube-system   kube-proxy-26sbr                     1/1     Running   0              5d8h   192.168.49.6    minikube-m05   <none>           <none>
kube-system   kube-proxy-2lg2j                     1/1     Running   0              5d8h   192.168.49.10   minikube-m09   <none>           <none>
kube-system   kube-proxy-4rh6n                     1/1     Running   0              5d8h   192.168.49.2    minikube       <none>           <none>
kube-system   kube-proxy-5n7qv                     1/1     Running   0              5d8h   192.168.49.8    minikube-m07   <none>           <none>
kube-system   kube-proxy-7s4dk                     1/1     Running   0              5d8h   192.168.49.3    minikube-m02   <none>           <none>
kube-system   kube-proxy-85klx                     1/1     Running   0              5d8h   192.168.49.4    minikube-m03   <none>           <none>
kube-system   kube-proxy-h4ppj                     1/1     Running   0              5d8h   192.168.49.9    minikube-m08   <none>           <none>
kube-system   kube-proxy-pb5td                     1/1     Running   0              5d8h   192.168.49.5    minikube-m04   <none>           <none>
kube-system   kube-proxy-qjc57                     1/1     Running   0              5d8h   192.168.49.7    minikube-m06   <none>           <none>
kube-system   kube-scheduler-minikube              1/1     Running   0              5d8h   192.168.49.2    minikube       <none>           <none>
kube-system   storage-provisioner                  1/1     Running   1 (5d8h ago)   5d8h   192.168.49.2    minikube       <none>           <none>
lsliwko@Mac kube-score-extender-service %
```

**FIGURE 4. Example output of *kubectl get pods* command used to list all pods across namespaces in the Minikube testbed. Similar listings were captured during evaluation to validate the correctness and quality of the resulting pod allocations.**

## VI. CONCLUSIONS

A core premise of this work is the reduction of cognitive overhead, which became immediately apparent during the configuration of the baseline experiments (as presented in Section IV.B). For instance, achieving the baseline behaviors in experimental Scenarios A and C required complex, multi-level YAML structures. These baselines relied on Kubernetes' default directives, such as *topologySpreadConstraints* and a combination of *podAffinity* and *podAntiAffinity*, respectively. Such configurations are verbose, difficult to author correctly, and prone to error. In contrast, the prototype required only a single allocation-hint annotation (e.g., "spread across zones"), demonstrating a significant and practical simplification in expressing user intent.

The development and evaluation of the semantic scheduling prototype revealed several key insights into integrating LLMs within latency-sensitive infrastructure systems. First, while LLMs demonstrated strong capability in translating natural-language hints into structured scheduling directives, their decoding reliability remains a central challenge. Even small misinterpretations of intents or metadata can lead to significant scheduling deviations. The need for deterministic, verifiable parsing underscores that LLMs should paired with strict schema validation, normalization layers, and human-readable audit trails to ensure operational safety in production environments.

In addition to these operational challenges, ethical and security issues must also be addressed. Algorithmic bias in LLM interpretations could result in unfair resource allocation. Because user hints are untrusted, strong protections against prompt injection attacks are required to safeguard cluster stability. Moreover, the risk of intentional resource exhaustion through crafted semantic requests underscores the need for robust admission controls.

The investigation has demonstrated that effective prompt design and schema constraints are as critical as model selection. Incremental improvements in prompt phrasing, output examples, and metadata formatting directly reduced false positives and extraction errors. This experience reinforced that LLM integration is as much an engineering discipline as a modeling challenge; robustness depends on careful pipeline design, not model capability alone.

The production deployment of the Score Extender involves operational challenges, specifically the complex configuration of *kube-scheduler* manifest files required to integrate custom extenders. Furthermore, because Kubernetes Control Plane nodes are typically isolated within secure namespaces, ensuring the implemented webhook endpoints are reachable from the control-plane node requires network policy management to maintain cluster security without obstructing scheduling communication.

The project also highlighted the importance of architectural separation between intent analysis and scheduling execution. The prototype exposed fundamental differences between research feasibility and production readiness: (i) embedding an LLM call within the synchronous */prioritize* path introduced unacceptable latency for real-world use, and (ii) the single-threaded Flask server, static scoring logic, and lack of fault tolerance are acceptable for evaluation but unsuitable for real workloads. Transitioning to a scalable, asynchronous service architecture and incorporating caching, concurrency, and adaptive fallback mechanisms will be essential for deployment in real clusters. The below list details the most crucial limitations in the current design:

- **A Single-threaded Python Flask Server**: While the prototype performed well during controlled evaluations, its Python Flask-based implementation is inherently single-threaded, making it unsuitable for production-scale workloads. Under concurrent scheduling requests, this architecture would quickly become a bottleneck, as each request must wait for the current one to complete before being processed. This serialization severely limits throughput and scalability, particularly in high-load cluster environments where multiple scheduling operations occur in parallel. A production-ready deployment would require migrating to a multi-threaded or asynchronous framework like FastAPI or aiohttp, and also to set up a load balancer to provide resilience.
- **Intent Classification Errors**: The scheduler's behavior depends critically on the correct identification of a presence of an intent. While top-tier models such as Amazon Nova Premier and Mistral Pixtral Large achieved high subset accuracy (97.45% and 95.86%, respectively), their performance is not flawless. The 9-18 aggregated false negatives represent cases where a user's hint (e.g., *spread_zones*) was completely missed, leading to violations of the intended scheduling policy.

TABLE V
EVALUATION SCENARIOS A-F

<table>
<tr><th></th><th>Baseline Kubernetes Manifest</th><th>Kubernetes Manifest with Allocation Hint</th><th>Objective / Metric / Results</th></tr>
<tr><td>Scenario A - Topology Spreading</td><td><pre>apiVersion: apps/v1
kind: Deployment
metadata: {name: test-scenario-a-baseline}
spec:
  replicas: 6
  selector: {matchLabels: {app: web}}
  template:
    metadata: {labels: {app: web}}
    spec:
      topologySpreadConstraints:
        - maxSkew: 1
          topologyKey: topology.kubernetes.io/zone
          whenUnsatisfiable: DoNotSchedule
          labelSelector:
            matchLabels:
              app: web
      containers: [{name: nginx, image: nginx}]</pre></td><td><pre>apiVersion: apps/v1
kind: Deployment
metadata: {name: test-scenario-a-hint}
spec:
  replicas: 6
  selector: {matchLabels: {app: web}}
  template:
    metadata:
      labels: {app: web}
      annotations:
        allocation_hint: |
          spread these pods evenly across all available
          zones for high availability
    spec:
      containers: [{name: nginx, image: nginx}]</pre></td><td>Objective: To evaluate the scheduler's ability to fulfill a high-availability policy.<br><br>Metric: The standard deviation of the pod count per zone. A lower value signifies a more even and superior distribution.<br><br>Result: Both, the baseline and intent-driven schedulers, achieved a perfect 3:3 split across the two zones, successfully fulfilling the spreading objective.</td></tr>
<tr><td>Scenario B - Resource Affinity[1]</td><td><pre>apiVersion: apps/v1
kind: Deployment
metadata: {name: test-scenario-b-baseline}
spec:
  replicas: 6
  selector: {matchLabels: {app: web}}
  template:
    metadata: {labels: {app: web}}
    spec:
      affinity:
        nodeAffinity:
          preferredDuringSchedulingIgnoredDuringExecution:
            - weight: 100
              preference:
                matchExpressions:
                  - key: hardware
                    operator: In
                    values:
                      - gpu
      containers: [{name: nginx, image: nginx}]</pre></td><td><pre>apiVersion: apps/v1
kind: Deployment
metadata: {name: test-scenario-b-hint}
spec:
  replicas: 6
  selector: {matchLabels: {app: web}}
  template:
    metadata:
      labels: {app: web}
      annotations:
        allocation_hint: |
          this is a critical ML training job, it must run
          on nodes with GPUs
    spec:
      containers: [{name: nginx, image: nginx}]</pre></td><td>Objective: To test the system's handling of resource preferences, which aligns with the prototype's implementation of the soft-affinity /prioritize verb.<br><br>Metric: The percentage of pods successfully placed on nodes labeled as having GPUs.<br><br>Result: Both, the baseline and intent-driven schedulers, placed 6 out of 6 pods (100%) on the minikube-m02 node. This demonstrates that it correctly interpreted the "must run" hint as a strong preference and its scoring logic.</td></tr>
<tr><td>Scenario C - Complex Co-location and Anti-Affinity[2]</td><td><pre>apiVersion: apps/v1
kind: Deployment
metadata: {name: test-scenario-c-baseline}
spec:
  replicas: 6
  selector: {matchLabels: {app: web}}
  template:
    metadata: {labels: {app: web}}
    spec:
      affinity:
        podAffinity:
          preferredDuringSchedulingIgnoredDuringExecution:
            - weight: 100
              podAffinityTerm:
                labelSelector:
                  matchExpressions:
                    - key: app
                      operator: In
                      values:
                        - database
                        - cache
                topologyKey: "topology.kubernetes.io/region"
        podAntiAffinity:
          requiredDuringSchedulingIgnoredDuringExecution:
            - labelSelector:
                matchLabels:
                  app: logging-agent
              topologyKey: "kubernetes.io/hostname"
containers: [{name: nginx, image: nginx}]</pre></td><td><pre>apiVersion: apps/v1
kind: Deployment
metadata: {name: test-scenario-c-hint}
spec:
  replicas: 6
  selector: {matchLabels: {app: web}}
  template:
    metadata:
      labels: {app: web}
      annotations:
        allocation_hint: |
          prefer to be in the same region as the
          'database' and 'cache' deployments, but avoid
          being on the same node as the 'logging-agent'
          pods.
    spec:
      containers: [{name: nginx, image: nginx}]</pre></td><td>Objective: To evaluate the system's expressiveness and its ability to handle multiple, mixed intents, which is expected to be a strength of the natural language approach.<br><br>Metric: Evaluate of whether the placement decisions successfully adhered to both the region affinity and node anti-affinity objectives.<br><br>Result (baseline): The baseline scheduler placed all 6 pods correctly, avoiding the logging-agent node (minikube-m02) and distributing them across minikube-m03, m06, m07, m08, and m09.<br><br>Result (intent-driven): The intent-driven scheduler also placed all 6 pods correctly, avoiding minikube-m02. It successfully fulfilled both objectives by placing all pods on minikube-m03 (near the cache deployment).</td></tr>
<tr><td>Scenario D - Rapid Burst Colocation</td><td><pre>apiVersion: apps/v1
kind: Deployment
metadata: {name: test-scenario-d-baseline}
spec:
  replicas: 20
  selector: {matchLabels: {app: web}}
  template:
    metadata: {labels: {app: web}}
    spec:
      affinity:
        podAffinity:
          requiredDuringSchedulingIgnoredDuringExecution:
            - labelSelector:
                matchLabels:
                  app: web
              topologyKey: "kubernetes.io/hostname"
      containers: [{name: nginx, image: nginx}]</pre></td><td><pre>apiVersion: apps/v1
kind: Deployment
metadata: {name: test-scenario-d-hint}
spec:
  replicas: 20
  selector: {matchLabels: {app: web}}
  template:
    metadata:
      labels: {app: web}
      annotations:
        allocation_hint: |
          Collocate all pods from this deployment on a
          single node.
    spec:
      containers: [{name: nginx, image: nginx}]</pre></td><td>Objective: To test the effectiveness of the recent placements cache in ensuring colocation logic works correctly during high-velocity pod creation bursts, compensating for main cache update latency.<br><br>Metric: The total number of distinct nodes used to place all replicas of the deployment. An ideal outcome is a single node.<br><br>Result: Both the baseline and intent-based schedulers successfully achieved the ideal outcome of a single distinct node, placing all 20 pods onto the same node.</td></tr>
<tr><td>Scenario E - Quantitative Resource Preference[3]</td><td><pre>apiVersion: apps/v1
kind: Deployment
metadata: {name: test-scenario-e-baseline}
spec:
  replicas: 6
  selector: {matchLabels: {app: web}}
  template:
    metadata: {labels: {app: web}}
    spec:
      affinity:
        nodeAffinity:
          preferredDuringSchedulingIgnoredDuringExecution:
            - weight: 100
              preference:
                matchExpressions:
                  - key: network-gbps
                    operator: Gt
                    values:
                      - "99"  # (>100 Gbps)
      containers: [{name: nginx, image: nginx}]</pre></td><td><pre>apiVersion: apps/v1
kind: Deployment
metadata: {name: test-scenario-e-hint}
spec:
  replicas: 6
  selector: {matchLabels: {app: web}}
  template:
    metadata:
      labels: {app: web}
      annotations:
        allocation_hint: |
          This is a high-bandwidth job, please place on
          nodes with at least 100Gbps network speed.
    spec:
      containers: [{name: nginx, image: nginx}]</pre></td><td>Objective: To evaluate the full pipeline, from the LLM's ability to parse specific numerical metadata (e.g., prefer_network_speed) to the scoring logic's ability to use that value correctly.<br><br>Metric: The count of pods successfully placed on nodes that are labeled in the testbed as having network speed of 100Gbps or more.<br><br>Result (baseline): The baseline scheduler placed only 1 of 6 pods on the preferred node. This demonstrates that the default Kubernetes scheduler's other plugins (like topology spreading) outweighed the soft nodeAffinity preference.<br><br>(Notably, a version of deployment manifest with requiredDuringSchedulingIgnoredDuringExecution works as expected.)<br><br>Result (intent-based): The intent-driven scheduler placed 6 of 6 pods on the correct node with network Gbps set to 100.</td></tr>
</table>

**Scenario F - Conflicting Intent Resolution**

```
apiVersion: apps/v1
kind: Deployment
metadata:
  name: test-scenario-f-baseline
spec:
  replicas: 2
  selector: {matchLabels: {app: web}}
  template:
    metadata:
      labels: {app: web}
    spec:
      affinity:
        podAffinity:
          # Hard constraint for colocation (from Scenario D)
          requiredDuringSchedulingIgnoredDuringExecution:
            - labelSelector:
                matchLabels:
                  app: web
              topologyKey: "kubernetes.io/hostname"
      topologySpreadConstraints:
        # Hard constraint for spreading (from Scenario A)
        - maxSkew: 1
          topologyKey: topology.kubernetes.io/zone
          whenUnsatisfiable: DoNotSchedule
          labelSelector:
            matchLabels:
              app: web
      containers: [{name: nginx, image: nginx}]
```

```
apiVersion: apps/v1
kind: Deployment
metadata:
  name: test-scenario-f-hint
spec:
  replicas: 2
  selector: {matchLabels: {app: web}}
  template:
    metadata:
      labels: {app: web}
      annotations:
        allocation_hint: |
          For high performance, collocate all pods on a
          single node. For high availability, you must
          also spread these pods across all zones.
    spec:
      containers: [{name: nginx, image: nginx}]
```

Objective: To test the behavior of the current simple additive scoring model when faced with directly conflicting user intents.

Metric: The baseline deployment manifest is configured with two contradictory hard constraints. The final placement distribution is observed to determine how the system prioritizes and resolves the trade-off between the colocation and spreading directives.

Result (baseline): The manifest utilizes *requiredDuringScheduling IgnoredDuringExecution* for both colocation and spreading. Once the first pod is assigned to minikube-m02, the second pod faces an irreconcilable conflict, i.e., it must be on minikube-m02 to satisfy colocation but is forbidden from that zone/node to satisfy spreading. This results in the second pod remaining in a Pending state and this is the expected outcome.

Result (intent-based): The test-scenario deployments manifest provides contradictory soft preferences; however, the prototype's additive scoring logic is designed to resolve such conflicts by summing the weighted scores. It finds a best-fit solution and schedules the pod. This is a known limitation of the prototype, which a future cost-function-based optimizer would address.

[1] The GPU count in Kubernetes is specified as an extended resource *nvidia.com/gpu* as a standard. The *nodeAffinity* mechanism (both *preferredDuringScheduling*... and *requiredDuringScheduling*...) only works with node labels, and not with node resources. Therefore, in order to simulate soft-affinity, a custom *hardware=gpu* label has been applied to the node *minikube-m02*.

[2] Scenario C included the deployment of pods from the *database* (on *minikube-m05*), *cache* (on *minikube-m03*), and *logging-agent* (on *minikube-m02*) deployments.

[3] Scenario E involved setting the network bandwidth to 10 Gbps on all Minikube nodes, except for node *minikube-m09*, where it was set to 100 Gbps to meet the test scenario requirements.

Conversely, the 10-17 aggregated false positives correspond to hallucinated intents, where the model inferred preferences not expressed by the user. Such misclassifications can result in unpredictable or suboptimal scheduling outcomes, as the system might apply unintended placement preferences or ignore critical user directives.

- **Metadata Extraction Errors**: Even when an intent is correctly identified, its associated metadata can be misinterpreted or extracted incorrectly. As shown in Table IV, the best-performing models on this metric, i.e., Claude 3 Sonnet (95.22%) and Amazon Nova Pro (94.40%), still exhibit nontrivial error rates. This limitation, observed repeatedly during prompt refinement, becomes particularly pronounced with complex list extractions. For instance, a hint such as "avoid us-east-1a and us-east-1b" might be erroneously parsed as ['us-east-1'], preventing the scheduler from excluding the specified zones. Similarly, an implicit hint like "high memory" may be incorrectly assigned a default numeric value (e.g., 1.0), rendering the resulting *prefer_memory_gb* intent ineffective. Such metadata errors can silently degrade scheduling accuracy, as the scoring logic depends on precise, structured values.
- **Unreliable Strength Interpretation**: The most significant decoding weakness, as highlighted in Table IV, is unreliable strength interpretation. The Overall Strength Accuracy metric measuring how well models map linguistic modifiers like "must" (strength = 1.5) or "prefer" (strength = 0.5) proved the most challenging. Even the best-performing models, Nova Premier and Mistral Pixtral Large, achieved only 71.79% accuracy, while the chosen Nova Pro model reached 67.85%. Consequently, in roughly one-third of cases, the model misinterprets the user's intended priority. A "critical" directive (strength 1.5) misclassified as a standard preference (strength 1.0) would cause the scheduler to treat it as a weak suggestion, allowing less important factors to dominate the decision. This systematic degradation of the intent strength undermines user expectations and highlights the need for explicit strength calibration or post-processing normalization in future iterations.

Overall, these lessons demonstrate that while LLM-driven intent recognition holds strong promise for advancing declarative scheduling, reliability, observability, and performance isolation must remain first-class design goals. Future iterations should focus on modularizing these components, enabling general-purpose, platform-agnostic scheduling augmentation built on stable, interpretable NLP foundations. The future works should cover the following:

- **Intent Analysis Prior to Scoring**: Although the prototype demonstrates the feasibility of translating natural-language hints into scheduling directives, its architecture introduces a significant performance bottleneck. The current design performs a synchronous LLM call within the */prioritize* handler, adding P95 latencies ranging from 0.87s to 5.34s, depending on the model. In production, this delay would be unacceptable during scheduling. A production-ready solution should move the LLM call outside the scheduling loop, likely most effectively to be implemented via a *MutatingAdmissionWebhook* that analyzes intents at pod creation, stores the structured JSON output in an annotation, and allows the scheduler extender to read it locally during */prioritize*. This would remove the synchronous dependency, eliminating latency and making semantic scheduling viable for high-throughput clusters.

- **Expanding Evaluation Dataset with Compound Multi-Objective Prompts**: The current 314-prompt evaluation dataset primarily covers cases translating into one or two intent objects. Future work should extend this dataset to include multi-level and conflicting scheduling intents, reflecting real-world scenarios where objectives such as co-location and fault tolerance interact. Incorporating such complex and contradictory prompts will better test the LLM's reasoning and consistency, highlighting the need for larger, context-aware models capable of resolving competing directives and handling richer semantic structures. This work has already commenced with the creation of the v2 dataset, available on the Model Hub [37] (file: schedule-intent-evaluation-v2.json), which introduces two new categories: Multi-Objective Optimization (focused on encoding multiple Intents) and Contradictory Prompt (providing opposing hints).
- **Pareto Frontier Optimization**: A key direction for future work is the integration of Pareto-based scoring into the scheduling framework. This approach would allow multiple, potentially conflicting objectives to be evaluated simultaneously, while preserving the dominance of higher-priority intents. Implementing a tiered logic scheme, where lower-strength objectives act as tie-breakers only when no higher-strength objective dominates, could improve decision-making in complex scheduling scenarios. Future iterations will explore the design and evaluation of this method.
- **Evaluation in a Parallel Execution Environment**: A notable limitation of the current prototype is its execution environment. While the Minikube testbed simulates a multi-node cluster, the core scoring logic within the Extender Service operates sequentially due to Flask's single-threaded nature. This means that even if multiple pod scheduling requests arrive concurrently at the extender, they are processed one after another. This sequential processing does not accurately reflect a production scenario where a busy scheduler extender handles numerous */prioritize* requests in parallel. Future work should involve reimplementing the extender logic using a multi-threaded server capable of parallel request handling. Evaluating the system under such concurrent load is crucial, as it will likely necessitate the introduction of sophisticated locking mechanisms. Specifically, concurrent scoring of pods belonging to the same deployment (e.g., during a scale-up) could lead to race conditions when updating or querying the recent placements cache or when calculating dynamic scores for *spread_* and *colocate_* intents. Therefore, implementing appropriate concurrency controls will be essential for ensuring the correctness and scalability of the intent-based scoring logic in a realistic, high-throughput environment.
- **General-Purpose Scheduling Grammar**: A key direction for future research is to decouple the core scheduling logic from its current tight integration with Kubernetes. In the prototype, Intent definitions and scoring logic depend on Kubernetes-specific topology labels (e.g., zone, rack) and its scheduler extender API. To generalize this concept, a universal scheduling grammar should be introduced - an abstract intermediary format for defining soft-affinity scheduling preferences. The LLM would translate natural-language hints into this grammar, while platform-specific adapters would convert it into concrete directives for systems such as Kubernetes, Mesos, or SLURM. This approach would decouple NLP-based intent recognition from any single platform, enabling a general-purpose framework for semantic workload orchestration across diverse environments.

In summary, this work successfully validates the core premise that LLMs can bridge the semantic gap in cluster scheduling, demonstrating high accuracy in translating human intent into structured scheduling policies. The prototype proves the concept of semantic soft affinity, shifting the bottleneck from human configuration error to the engineering challenges of integration, specifically regarding synchronous latency and reliable metadata decoding. This research thus confirms the viability of intent-driven orchestration while clearly outlining the architectural and reliability engineering required to transition such a system from a novel prototype to a production-grade infrastructure component.

## ACKNOWLEDGMENTS

The authors wish to express their sincere gratitude to Standard Chartered Bank for its material support, which contributed significantly to the research and investigation.

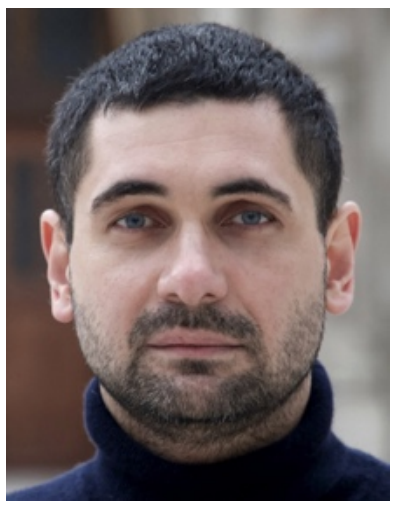

**LESZEK SLIWKO** was awarded a PhD in Parallel and Distributed Computing from the University of Westminster (London, UK) in 2019 for his work on developing a novel AI-driven load balancer for cloud systems. He has presented at international events such as IEEE conferences, Machine Intelligence Research Lab lectures, and popular seminars like Scala in the City, JVM Roundabout, and Scala Central. His talks are available at https://www.youtube.com/@lsliwko. He is keen on experimenting with the latest technologies, particularly those related to High-Performance Computing and Machine Learning. Leszek has been an IEEE Professional member since 2019. A brief description of his selected projects can be found at https://meta-analyzer.com/projects/.

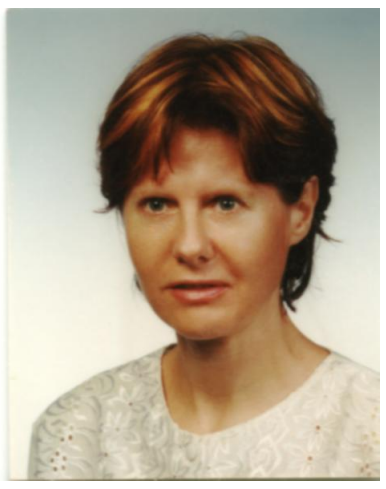

**JOLANTA MIZERA-PIETRASZKO** received the M.Sc. degree in computer science from the University of Economics and Ph.D. degree in computer science from Wrocław University of Science and Technology, Poland, in 2014. She is with Opole University of Technology, Poland. Her research interests focus on multilingual natural language processing, medical informatics (she has been Head of Medical Informatics Research Group), computational linguistics, data and text mining, web systems, information retrieval, automated translation, cybersecurity in military systems. She serves on ACM Europe Technology Policy Committee in Differential AI and Data Science. She authored and co-authored more than 100 papers in conference proceedings and journals including an innovative MT model published by Patent Office of the Republic of Poland. She has chaired and served on TPC of around 80 international conferences. She serves as Associate Editor for IEEE Access, Featured Reviewer for ACM Computing Reviews, Editorial Advisory Board member for Recent Advances in Computer Science and Communications and a Guest Editor for some ISI journals like Big Data Research, Computing Journal, and the Journal of Universal Computer Science. She is an Expert for European Commission in evaluation of research projects on AI under HORIZON and some national Polish research agencies. She is an ACM Professional Member, British Computer Society, Information Retrieval Specialist Group and IEEE ComSoc Member.